\documentclass{article}

\usepackage{microtype}
\usepackage{graphicx}
\usepackage{subcaption}
\usepackage{booktabs}

\usepackage{kotex}
\usepackage{array}
\usepackage{multirow}
\usepackage{xcolor, color, colortbl}
\definecolor{Gray}{gray}{0.92}
\usepackage{wrapfig}

\newcolumntype{C}[1]{>{\centering\arraybackslash}p{#1}}
\newcolumntype{L}[1]{>{\arraybackslash}p{#1}}

\usepackage{tablefootnote, footnote}
\usepackage[bottom]{footmisc}
\usepackage{amsmath, amsfonts}
\usepackage{tabulary}
\usepackage{bbm}
\usepackage{enumitem}

\usepackage{hyperref}

\usepackage[accepted]{icml2023}

\usepackage{amsmath}
\usepackage{amssymb}
\usepackage{mathtools}
\usepackage{amsthm}
\usepackage[capitalize,noabbrev]{cleveref}

\theoremstyle{plain}

\theoremstyle{definition}

\theoremstyle{remark}

\usepackage[textsize=tiny]{todonotes}
\icmltitlerunning{End-to-End Multi-Object Detection with a Regularized Mixture Model}

\begin{document}

\twocolumn[
\icmltitle{End-to-End Multi-Object Detection with a Regularized Mixture Model}

\icmlsetsymbol{equal}{*}

\begin{icmlauthorlist}
\icmlauthor{Jaeyoung Yoo}{equal,wt}
\icmlauthor{Hojun Lee}{equal,snu-dii}
\icmlauthor{Seunghyeon Seo}{snu-ipai}
\icmlauthor{Inseop Chung}{snu-dii}
\icmlauthor{Nojun Kwak}{snu-dii,snu-ipai}
\end{icmlauthorlist}

\icmlaffiliation{wt}{NAVER WEBTOON AI}
\icmlaffiliation{snu-dii}{Department of Intelligence and Information Science, Seoul National University}
\icmlaffiliation{snu-ipai}{Interdisciplinary Program in Artificial Intelligence, Seoul National University}
\icmlcorrespondingauthor{Jaeyoung Yoo}{yoojy31@webtoonscorp.com}
\icmlcorrespondingauthor{Hojun Lee}{hojun815@snu.ac.kr}
\icmlcorrespondingauthor{Nojun Kwak}{nojunk@snu.ac.kr}

\icmlkeywords{Machine Learning, ICML}

\vskip 0.3in
]

\printAffiliationsAndNotice{\icmlEqualContribution}

\begin{abstract}
Recent end-to-end multi-object detectors simplify the inference pipeline by removing hand-crafted processes such as non-maximum suppression (NMS). However, during training, they still heavily rely on heuristics and hand-crafted processes which deteriorate the reliability of the predicted confidence score. In this paper, we propose a novel framework to train an end-to-end multi-object detector consisting of only two terms: negative log-likelihood (NLL) and a regularization term. In doing so, the multi-object detection problem is treated as density estimation of the ground truth bounding boxes utilizing a regularized mixture density model. The proposed \textit{end-to-end multi-object Detection with a Regularized Mixture Model} (D-RMM) is trained by minimizing the NLL with the proposed regularization term, maximum component maximization (MCM) loss, preventing duplicate predictions. Our method reduces the heuristics of the training process and improves the reliability of the predicted confidence score. Moreover, our D-RMM outperforms the previous end-to-end detectors on MS COCO dataset. \textsc{Code will be available.}
\end{abstract}

\section{Introduction}
\label{sec:intro}
``How can we train a detector to learn a variable number of ground truth bounding boxes for an input image without duplicate predictions?'' This is a fundamental question in the training of end-to-end multi-object detectors \cite{carion2020detr}. In multi-object detection, each training image has a different number of bounding box coordinates and the corresponding class labels. Thus, the network output is challenging to match one-to-one with the ground truth. Conventional methods \cite{ren2015fasterRCNN,liu2016ssd,redmon2016yolo} train the detector by assigning a ground truth to many duplicate predictions, but they cannot directly obtain the final predictions without relying on non-maximum suppression (NMS) in the inference phase.

\begin{figure*}[t]
    \begin{center}
    \includegraphics[width=0.88\linewidth]{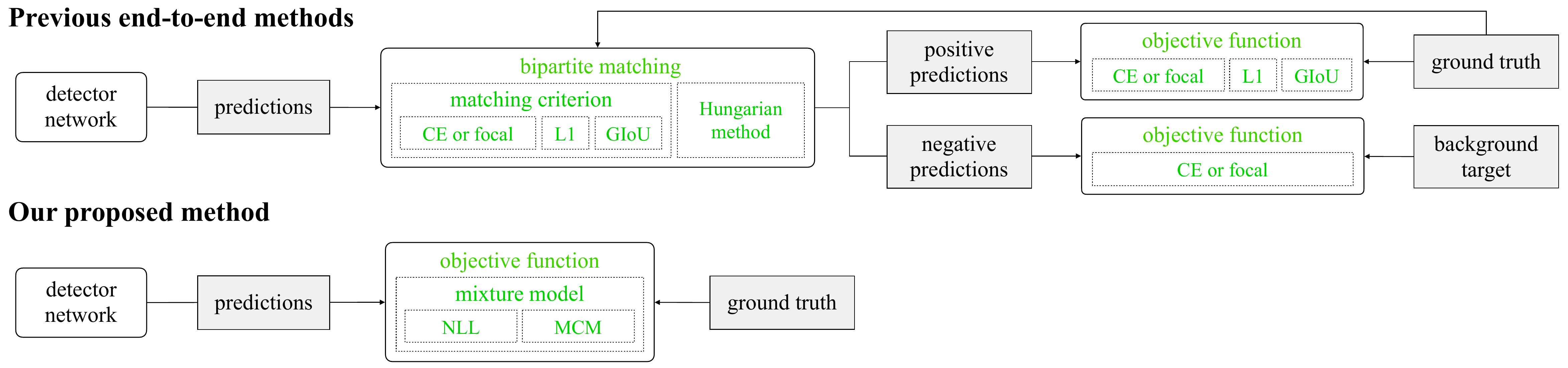}
    \end{center}
    \vspace{-3.7mm}
    \caption{Training pipeline of end-to-end multi-object detectors. First row : previous end-to-end method such as \cite{carion2020detr,sun2021sparse}. Second row: our proposed method. The green text denotes the manually designed training components.}
    \label{fig:illustraion}
    \vspace{-1.6mm}
\end{figure*}

As an answer to the opening question, recent end-to-end multi-object detection methods address the training of detector by searching for unique assignments between the predictions and the ground truth via bipartite matching (Figure \ref{fig:illustraion}). The assigned positive prediction by bipartite matching learns the corresponding ground truth by a pre-designed objective function. On the other hand, the unassigned negative predictions do not care about ground truth information but are trained as backgrounds. Unlike conventional detectors, end-to-end methods can obtain final predictions from detector networks through bipartite matching-based training without relying on NMS in the inference phase.

\begin{figure}
    \centering
    \includegraphics[width=0.77\linewidth]{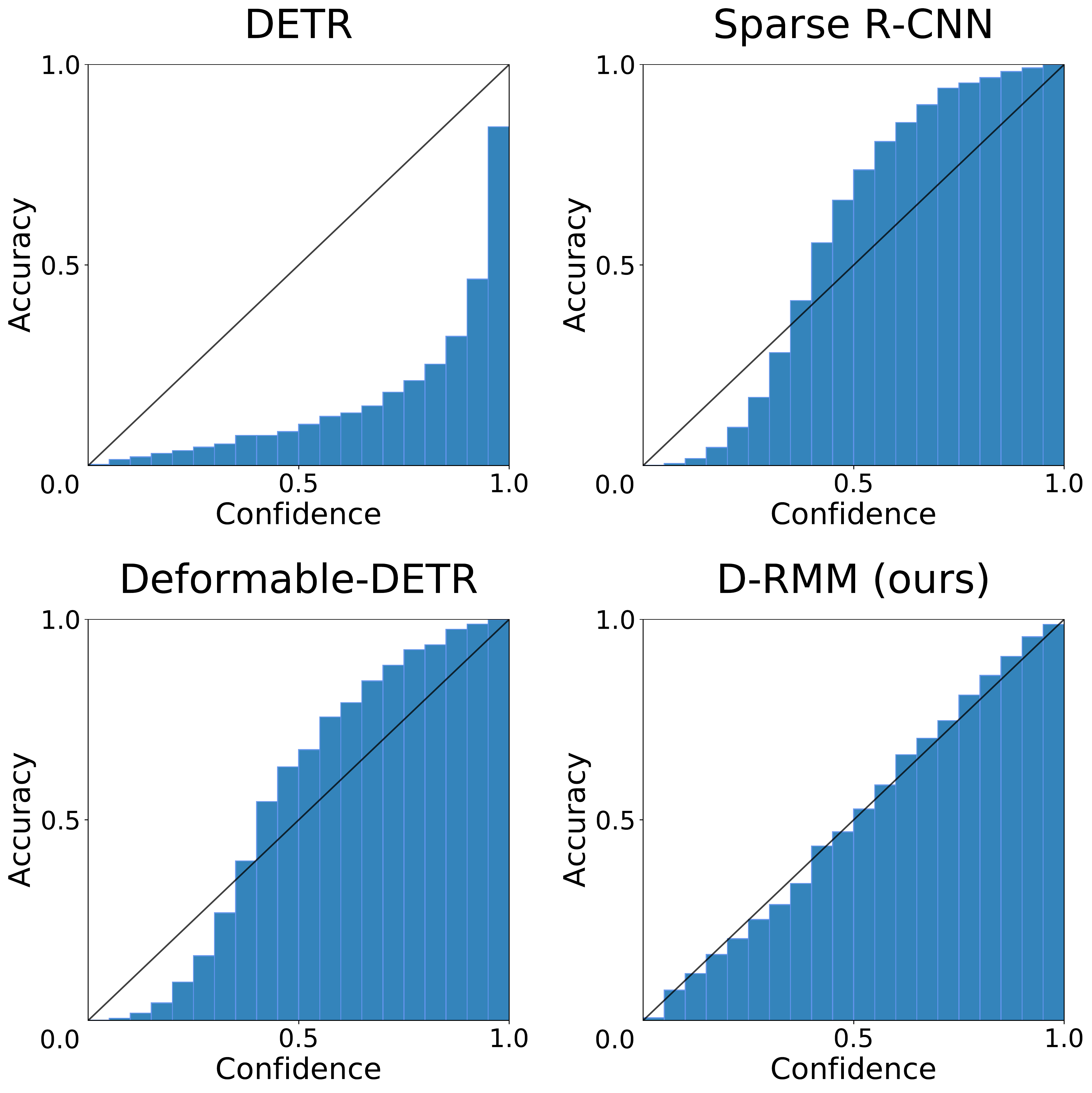}
    \vspace{-3.0mm}
    \caption{Confidence score \textit{vs.} accuracy of various end-to-end detectors. Accuracy is measured for the predicted bounding boxes with the confidence score in the corresponding confidence bin.} 
    \vspace{-3.0mm}
    \label{fig:calibration}
\end{figure}

However, the training of the current end-to-end multi-object detectors has several drawbacks as follows: 
\vspace{0.4mm}
\\
\textbf{Immoderate heuristics.} \quad In the training process, the current end-to-end methods use heuristically designed objective functions and  matching criteria between the ground truth and a prediction. For instance, DETR, the representative end-to-end method, uses the following combination of losses as the training objective and matching criterion:
\begin{equation}
\begin{split}
\mathcal{L} &= w_1 \cdot \mathcal{L}_{CE}  + w_2\cdot \mathcal{L}_{L1} + w_3\cdot \mathcal{L}_{GIoU},
\end{split}
\label{eq:matcher}
\end{equation}
where, $\mathcal{L}_{CE}$, $\mathcal{L}_{L1}$, and $\mathcal{L}_{GIoU}$ are cross-entropy, L1 and GIoU \cite{rezatofighi2019generalized} loss with its balancing hyper-parameters ($w_1$, $w_2$, and $w_3$) respectively. Deformable DETR \cite{zhu2020deformable} and Sparse R-CNN \cite{sun2021sparse}, other popular end-to-end methods, replace the cross-entropy loss with the focal loss \cite{lin2017focal}. 
\vspace{0.4mm}
\\
\textbf{Hand-crafted assignment.} \quad Since there are many possible pairs between ground truths and predictions, the end-to-end methods need to find an optimal set of pairs among them. To solve this assignment problem, most end-to-end detectors utilize hand-crafted algorithm such as the Hungarian method \cite{kuhn1955hungarian} in the training pipeline to find a good bipartite matching.
\vspace{0.4mm}
\\
\textbf{Unreliable confidence.} \quad In the aspect of the nature of human recognition, the confidence score of the prediction is regarded as an estimate of the accuracy \cite{guo2017oncalibration}. For the confidence score to be reliable, which means that the confidence score is to be used directly as an accuracy estimate, it should have a probabilistic meaning. However, in the training process based on bipartite matching, the predictions are discretely classified as either positive or negative by a hand-crafted assignment process and a heuristic matching criterion rather than from a probabilistic point of view. In addition, the weight in the focal loss, which is a general loss term to learn class probability, is controlled by hyper-parameters and does not provide a clear probabilistic basis. These heuristics and hand-designed training processes lead to gaps between predicted confidence scores and actual accuracy. Figure \ref{fig:calibration} shows the difference between the confidence score and the actual accuracy in the prediction of previous end-to-end detectors (DETR, Deformable DETR, Sparse R-CNN) and our probabilistic model (D-RMM).

In this study, we aim to overcome the aforementioned limitations and answer the opening question better. To this end, we propose a novel end-to-end multi-object detection framework, D-RMM, where we reformulate the end-to-end multi-object detection problem as a parametric density estimation problem. Our detector estimates the distribution of bounding boxes and object class using a mixture model. The proposed training loss function consists of the following two terms: the negative log-likelihood (NLL) and the maximum component maximization (MCM) loss. The NLL loss is a simple density estimation term of a mixture model. The MCM loss is the regularization term of the mixture model to achieve non-duplicate predictions. The NLL loss is calculated without any matching process, and the MCM loss only uses a simple maximum operation as matching. 
The contributions of our study are summarized as follows:

\begin{itemize}[itemsep=1ex, leftmargin=+5mm]
\vspace{-2mm}
\item We approach the end-to-end multi-object detection as a mixture model-based density estimation. To this end, we introduce an intuitive training objective function and the corresponding network architecture. 
\vspace{-2mm}
\item We replace the heuristic objective function consisting of several losses with a simple negative log-likelihood (NLL) and the regularization (MCM) terms of a mixture model from a probabilistic point of view.
\vspace{-2mm}
\item Thanks to the simplicity of the NLL and MCM loss, they are directly calculated from the network outputs without any additional process, such as bipartite matching.
\vspace{-2mm}
\item As can be seen in Figure \ref{fig:calibration}, the predictions of our method provide more reliable confidence scores.
\vspace{-2mm}
\item Our work outperforms the structural baselines (Sparse R-CNN, AdaMixer) and other state-of-the-art end-to-end multi-object detectors on MS COCO dataset.
\end{itemize}

\section{Related Works}
Most modern deep-learning-based object detectors require post-processing to remove redundant predictions (e.g. NMS) from dense candidates in estimating final bounding boxes \cite{redmon2016yolo,liu2016ssd,ren2015fasterRCNN}. Instead of depending on manually-designed post-processing, a line of recent works \cite{hu2018relation,carion2020detr,sun2021sparse} has proposed end-to-end object detection methods which output final bounding boxes directly without any post-processing in both the training and inference phase.

Recently, end-to-end methods \cite{hu2018relation,carion2020detr,zhu2020deformable} that do not use NMS-based post-processing have been proposed. DETR \cite{carion2020detr} proposes the training process for end-to-end detectors using the Hungarian algorithm \cite{kuhn1955hungarian}, which yields an optimal bipartite matching between $N \times K$ samples. This training process has become a standard for end-to-end detectors. Among them, Sparse R-CNN \cite{sun2021sparse} is one of the representative methods in which a fixed set of learned proposal boxes and features are used. Sparse R-CNN argues that it has a simpler framework than other end-to-end detectors \cite{carion2020detr,zhu2020deformable}, but it still uses Hungarian-algorithm-based bipartite matching for training.

However, the training of end-to-end detectors based on bipartite matching relies on heuristic objective functions, matching criteria, and hand-assigned algorithms. It is also known that the efficiency and stability of training are impaired due to the limited supervision by bipartite matching. \cite{jia2022detrs, li2022dn}

Another line of research has focused on removing the heuristics of the ground truth assignment process. Among them, Mixture Density Object Detector (MDOD) \cite{Yoo_2021_ICCV} reformulated the multi-object detection task as a density estimation problem of bounding box distributions with a mixture model. This enabled MDOD to perform regression without an explicit matching process with ground truths. However, MDOD still requires the matching process for training a classification task. Furthermore, it is not an end-to-end method and cannot replace the training process based on bipartite matching.

In this paper, we extend the density-estimation-based multi-object detector to an end-to-end method that does not need the deduplication process for the predictions. In addition, our D-RMM is trained as an end-to-end detector by directly calculating the loss from the network outputs, unlike other end-to-end methods that rely on an additional process such as the Hungarian method for bipartite matching. Our work greatly simplifies the training process of the end-to-end multi-object detector by using a straightforward strategy.

\section{The D-RMM Framework}
\subsection{Mixture model} \label{sec:method_mixture_model}
For the multiple ground truths $g = \{g_1, ..., g_{N}\}$ on an image $X$, each ground truth $g_i$ contains the coordinates of an object's location $b_i = \{b_{i,l},b_{i,t},b_{i,r},b_{i,b}\}$ (left, top, right, and bottom) and a one-hot class information $c_i$. The D-RMM network conditionally estimates the distribution of the $g$ for an image $X$ using a mixture model. 

The mixture model consists of two types of probability distribution: Cauchy (continuous) for bounding box coordinates and categorical (discrete) for class estimation. The Cauchy distribution is a continuous probability distribution that has a shape similar to the Gaussian distribution. However, it has heavier tails than the Gaussian, and is known to be less likely to incur underflow problems due to floating-point precision \cite{Yoo_2021_ICCV}. We use the 4-dimensional Cauchy to represent the distribution of the object's location coordinates. Also, a categorical distribution is used to estimate the object's class probabilities for the one-hot class representation. The probability density function of our mixture model is defined as follows:
\vspace{-1mm}
\begin{equation} \label{eq:mog}
\begin{aligned}
p(g_{i}|X) &= \sum_{k}^{K} \pi_{k}\mathcal{F}(b_i; \mu_k, \gamma_k) \mathcal{P}(c_i; p_k).
\end{aligned}
\end{equation}
Here, the $k$ is the index for the $K$ mixture components and the corresponding mixing coefficient is denoted by $\pi_k$. $\mathcal{F}$ and $\mathcal{P}$ denote the probability density function of the Cauchy and the probability mass function of the categorical distribution\, respectively. The parameters $\mu_k = \{\mu_{k,l}, \mu_{k,t}, \mu_{k,r}, \mu_{k,b}\}$ and $\gamma_k = \{\gamma_{k,l}, \gamma_{k,t}, \gamma_{k,r}, \gamma_{k,b}\}$ are the location and scale parameters of a Cauchy distribution, while $p_k = \{p_1, ..., p_C\}$ is the class probability of a categorical distribution. Here, $C$ is the number of possible classes for an object excluding the background class. To avoid over-complicating the mixture model, each element of $b_i$ is assumed to be independent of others. Thus, the probability density function of the Cauchy is factorized as
\begin{equation} \label{eq:gaussian}
\begin{aligned}
\mathcal{F} (b_i; \mu_k, \gamma_k) &= \prod_{j \in D} \mathcal{F}(b_{i,j}; \mu_{k,j}, \gamma_{k,j}), \quad D = \{l, t, r, b\}.
\end{aligned}
\vspace{-3mm}
\end{equation}

\subsection{Architecture}
For the implementation of our D-RMM, we adopt the overall architecture of Sparse R-CNN \cite{sun2021sparse} and its network characteristics such as learnable proposal box, dynamic head, and iteration structure due to the intuitive structure and fast training compared to the DETR \cite{carion2020detr} and Deformable DETR \cite{zhu2020deformable}. Also, we applied D-RMM to AdaMixer \cite{gao2022adamixer} while maintaining its structural characteristics.

\begin{figure}[t]
\begin{center}
\includegraphics[width=0.95\linewidth]{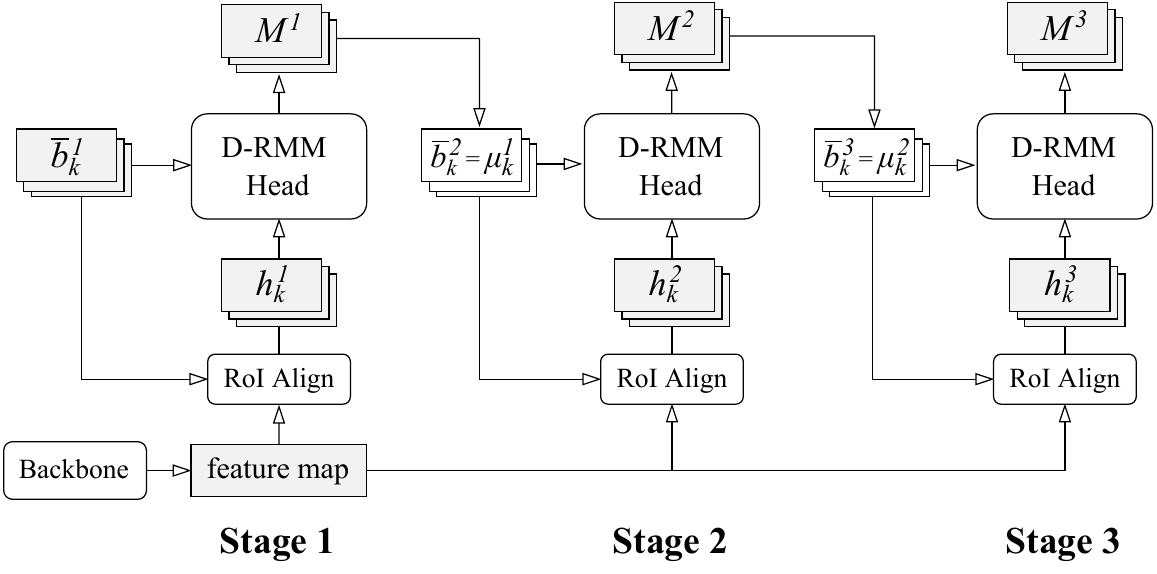}
\end{center}
\vspace{-5.0mm}
\caption{3-stage example of D-RMM architecture. D-RMM head predicts the mixture model's parameters $M^s$ from the proposal boxes $\bar{b}_k^s$ and an input image $X$.}
\label{fig:architecture}
\vspace{-1.5mm}
\end{figure}

Figure \ref{fig:architecture} shows the overview of our D-RMM network when the 3-stage iteration structure is used. First, the backbone network outputs the feature map from the input image $X$. In the first stage, a set of RoI features $h^1 = \{h_1^1, ..., h_K^1\}$ is obtained through RoI align process from the predefined learnable proposal boxes $\bar{b}^1 = \{\bar{b}_1^1, ..., \bar{b}_K^1\}$ and the feature map. Then, D-RMM head predicts $M^1 = \{\pi^1, \mu^1, \gamma^1, p^1, o^1\}$, the parameters of the mixture model ($\pi^1, \mu^1, \gamma^1, p^1$) and the objectness score ($o^1$), from $h^1$. Here, the number of mixture components $K$ equals the number of proposal boxes. In the $s$-th stage ($s \ge 2$), the process from RoI align to D-RMM head is repeated. $\mu^{s-1} \in \mathbb{R}^{4}$ which is the predicted location vector in the previous stage, is used as the proposal boxes $\bar{b}^s$ for the current stage $s$. Following \cite{sun2021sparse}, we adopt the 6-stage iteration structure.

\begin{figure}[t]
\begin{center}
\includegraphics[width=0.87\linewidth]{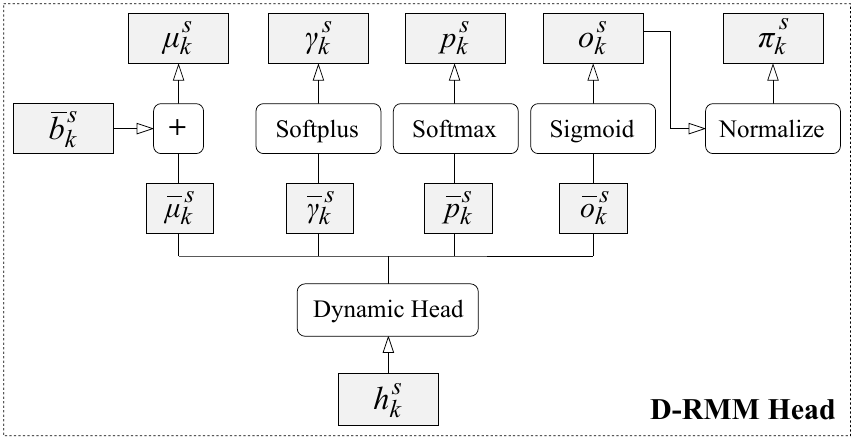}
\end{center}
\vspace{-4.5mm}
\caption{Structure of D-RMM head. D-RMM head predicts the parameters of the mixture model ($\pi_k^s$, $\mu_k^s$ $\gamma_k^s$ and $p_k^s$) and the objectness score ($o_k^s$) from a proposal box $\bar{b}_k^s$ and a RoI feature $h_k^s$.}
\label{fig:head}
\vspace{-3mm}
\end{figure}

The details of D-RMM head are illustrated in Figure \ref{fig:head}. The dynamic head outputs $\bar{\mu}_k^s, \bar{\gamma}_k^s, \bar{p}_k^s$ and $\bar{o}_k^s$ from $h_k^s$. The location parameter $\mu_k^s \in \mathbb{R}^{4}$ represents the coordinates of a mixture component and is produced by adding $\bar{b}_k^s$ to $\bar{\mu}_k^s$. The positive scale parameter $\gamma_k^s \in \mathbb{R}^{4}$ is obtained by applying the softplus activation \cite{dugas2000softplus} that always converts $\bar{\gamma}_k^s$ into a positive value. The object class probability $p_k^s \in \mathbb{R}^{C}$ is calculated by applying softmax function to $\bar{p}_k^s$ along the class dimension.

Note that, the probability of whether it is an object or not is not computed through $p_k^s$ but computed using an alternative way we propose to learn objectness score. Returning to the nature of the probability distribution, we utilize the properties of the mixture model. In the mixture model, the probability of a mixture component is expressed as a mixture coefficient $\pi_k^s$. In other words, the mixture component that is likely to belong to an object area has a higher $\pi_k^s$ value. In this aspect, we assume that $\pi$ could be regarded as the scaled objectness score such that $\sum_{k}^{K} \pi_k^s$ equals 1. From this assumption, we propose to express the mixture coefficient $\pi$ using the objectness score $o$. As shown in Figure \ref{fig:head}, the sigmoid activation outputs $o_k^s$ from $\bar{o}_k^s$. And then, $\pi_k^s$ is calculated by normalizing $o_k^s$ as 
$\pi_k^s = \frac{o_k^s}{ \sum_{k'=1}^{K} o_{k'}^s}$.

\begin{figure}[t]
\begin{center}
\includegraphics[width=0.92\linewidth]{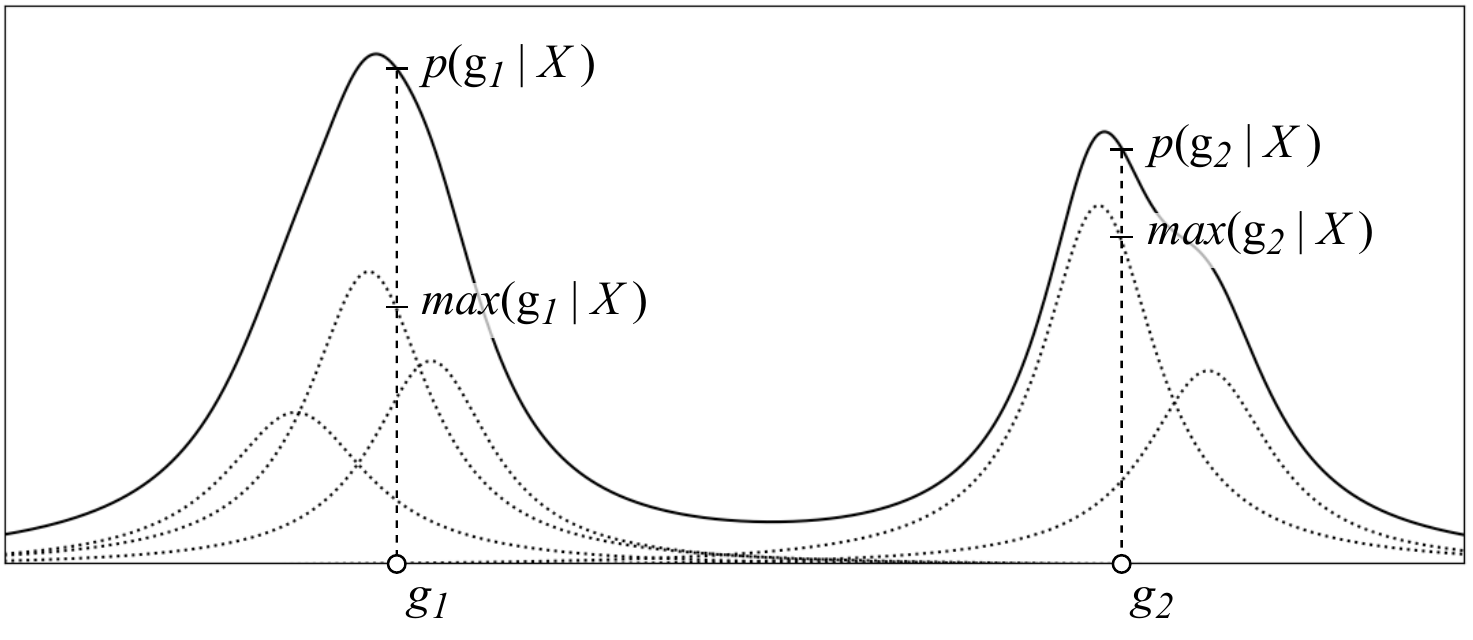}
\end{center}
\vspace{-6mm}
\caption{Illustration of 1-D example for the MCM loss ($\mathcal{L}_{MCM}$). By reducing the difference between $max(g_i|X)$ and  $p(g_i|X)$, it is trained to require only one mixture component to represent one $g_i$. Hence it restrains from having multiple components to represent the same single $g_i$.}
\vspace{-3mm}
\label{fig:mcm}
\end{figure}

\subsection{Training} \label{sec:training}
The D-RMM network is trained to maximize the likelihood of $g$ for the input image $X$ through the mixture model. The loss function is simply defined as the negative log-likelihood (NLL) of the probability density function as follows:
\begin{align}\label{eq:loss_nll}
    \mathcal{L}_{NLL} &= -\log p(g_i|X) \\&= -\log \sum_{k}^{K} \pi_k^s\mathcal{F}(b_i; \mu_k^s, \gamma_k^s) \mathcal{P}(c_i; p_k^s)
\end{align}
The D-RMM network learns the coordinates of the bounding box and the probability of the object class as $\mu$ and $p$ by minimizing the NLL loss ($\mathcal{L}_{NLL}$). The mixture coefficient $\pi$ learns the probability of a mixture component $\mathcal{F}(b_i) \mathcal{P}(c_i)$ that represents the joint probability for both box coordinates and object class. The objectness score $o$ is not directly used to calculate the NLL loss, but it is trained through $\pi$ (see Figure \ref{fig:head}). 

Here, we need to consider that the NLL loss does not restrict the distributional redundancy between multiple mixture components for single ground truth. This problem could lead to duplication of the predicted bounding boxes, as well as dispersion of the probability for one object to several mixture components. Thus, we introduce the maximum component maximization (MCM) loss which is the regularization term to the density estimation of the mixture model:
\begin{align}\label{eq:loss_lms}
\mathcal{L}_{MCM} &= -\log \frac {\max(g_i|X)} {p(g_i|X)}  \\ 
&= -\log \frac {\max(g_i|X)} {\sum_{k=1}^K \pi_k^s\mathcal{F}(b_i;\mu_k^s,\gamma_k^s)\mathcal{P}(c_i;p_k^s)},     
\end{align}
\begin{equation} \label{eq:max_likelihood}
\begin{aligned}
\max(g_i|X) &= \max_{k \in \{1, ..., K\}}(\pi_k^s\mathcal{F}(b_i;\mu_k^s,\gamma_k^s)\mathcal{P}(c_i;p_k^s))
\end{aligned}
\end{equation}
Figure \ref{fig:mcm} shows 1-D example for the MCM loss ($\mathcal{L}_{MCM}$). Minimizing the MCM loss reduces the difference of likelihood between $max(g_i|X)$ and $p(g_i|X)$. Through this, the mixture model is trained to maximize the probability of only one mixture component for one ground truth while reducing the probability of other adjacent components.
The total loss function is defined as follows: $\mathcal{L} = \mathcal{L}_{NLL} + \beta \times \mathcal{L}_{MCM}$, where $\beta$ balances between the NLL and the MCM loss. The total loss ($\mathcal{L}$) is computed for all stages of D-RMM, then summed together and back-propagated. To calculate the total loss, we do not need any additional process such as bipartite matching.

\subsection{Inference}
In the inference, $\mu$ of the last stage is used as the coordinates of the predicted bounding boxes. The class probability $p_k \in \mathbb{R}^C$ of $k$-th mixture component is the softmax output but, just the probability for the class of an object without background probability. Thus, we do not directly use $p$ as a confidence score of our prediction. Instead, the objectness score $o$ learned though the mixing coefficient $\pi$ is used with $p$. The confidence score of an $k$-th output prediction for class $c$ is calculated as $p_{k,c} \times o_k$ where $p_{k,c}$ is the $c$-th element of $p_k$. In the same manner as other end-to-end multi-object detectors, D-RMM also obtains final predictions without any duplicate bounding box removal process such as NMS.

\section{Experiments}
\subsection{Experimental details}
\noindent \textbf{Dataset.} \quad We evaluate D-RMM on MS COCO 2017 \cite{lin2014microsoft}.
Following the common practice, we split the dataset into 118K images for the training set, 5K for the validation set, and 20K for the test-dev set. We adopt the standard COCO AP (Average Precision) and AR (Average Recall) at most 100 top-scoring detections per image as the evaluation metrics. We report analysis results and comparison with a baseline on the validation set and compare with other methods on the test-dev and validation set.

\vspace{0mm}
\noindent \textbf{Training.} \quad As mentioned in Section \ref{sec:intro} and \ref{sec:method_mixture_model}, we model bounding box coordinates as Cauchy distributions and class probability as categorical distributions. We applied D-RMM to Sparse R-CNN (S-RCNN) and AdaMixer. For analysis, we adopt Sparse R-CNN architecture with 300 proposals. Unless specified, we followed the hyper-parameters of each original paper. As backbones, ResNet50 (R50), ResNet101 (R101) \cite{he2016resnet} and Swin Transformer-Tiny (Swin-T) \cite{liu2021swin} with Feature Pyramid Network (FPN) \cite{lin2017feature} are adopted, which are pretrained on ImageNet-1K \cite{imagenet}. The parameter $\beta$ for balancing the losses $\mathcal{L}_{NLL}$ and $\mathcal{L}_{MCM}$ is set to 0.5. $\beta$ was found experimentally, and related experiments are in Appendix Section \ref{appendix:sec:mcm_beta}. Synchronized batch normalization \cite{peng2018megdet} is applied for consistent learning behavior regardless of the number of GPUs. More details like batch size, data augmentation, optimizer and training schedule for reproducibility are in Appendix \ref{appendix:sec:training_details}.

\vspace{0mm}
\noindent \textbf{Inference.} \quad We select top-100 bounding boxes among the last head output according to their confidence scores without any further post-processing, such as NMS. 

\subsection{Comparison with Sparse R-CNN} 
\label{sec:comparison_and_ablation}
Table \ref{tab:comparison_with_srcnn} presents a comparison between Sparse R-CNN and D-RMM with different backbone networks on COCO validation set. We achieve significant AP improvement over Sparse R-CNN while maintaining the FPS on a similar level. There exists a slight gain ($\leq$ 0.1 FPS) in inference speed due to implementation details. The analysis of performance improvement is as follows.

\begin{table}[t]
\vspace{-2mm}
\caption{Comparison with Sparse R-CNN \cite{sun2021sparse}. FPS is measured as a network inference time excluding data loading on a single NVIDIA TITAN RTX using MMDet \cite{mmdetection} with batch size 1.}
\vspace{2.54mm}
\centering
\footnotesize
\renewcommand*{\arraystretch}{1.10}
\begin{tabular}{c|c|ccc|c}
\hline
Method & Backbone & AP & AP$_{50}$ & AP$_{75}$ & FPS\\
\hline\hline
S-RCNN & \multirow{2}{*}{R50 FPN}  &  45.0  & 64.1   & 49.0    & 22.7 \\ %
D-RMM &  &  47.0  &  64.8  &  51.6   & 22.8 \\ %
\hline
S-RCNN & \multirow{2}{*}{R101 FPN}  &  46.4  & 65.6  & 50.7 & 17.3 \\ %
D-RMM & &  48.0  &  65.7  &  52.6   & 17.3 \\ %
\hline
S-RCNN & \multirow{2}{*}{Swin-T FPN}  &  47.4  & 67.1  & 52.0 & 16.4 \\ %
D-RMM &  & 49.9   &  68.1  &  55.1 & 16.5 \\ %
\hline
\end{tabular}
\label{tab:comparison_with_srcnn}
\vspace{-3mm}
\end{table}

\begin{table}[t]
\caption{$B$ denotes Bipartite Matching, $M$ denotes Max. Both the matching cost and objective function of the 2nd row are the NLL loss. And in the case of the objective function, the NLL loss is computed only for matched predictions. In other words, unlike the 3rd-5th rows, the 2nd row is not a mixture model.}
\vspace{2.0mm}
\footnotesize
\renewcommand*{\arraystretch}{1.10}
\begin{center}
\begin{tabular}{l|cccc|c}
\hline
Method & REG & CLS & NLL & MCM & AP  \\
\hline\hline
S-RCNN & $\checkmark (B)$ & $\checkmark (B)$ &                  &                  & 45.0 \\
       &                  &                  & $\checkmark (B)$ &                  & 46.3 \\
       &                  &                  & $\checkmark $    &                  & 35.6 \\ 
       &                  &                  & $\checkmark$     & $\checkmark (B)$ & 46.9 \\
D-RMM  &                  &                  & $\checkmark$     & $\checkmark (M)$ & 47.0 \\
\hline
\end{tabular}
\label{tab:from_srcnn_to_ours}
\vspace{-3mm}
\end{center}
\end{table}

\noindent \textbf{Objective function.} \quad In Table \ref{tab:from_srcnn_to_ours}, we compared the APs by varying the objective function and observed the following. First, in Sparse R-CNN, modeling the objective function and the matching cost jointly with the NLL loss is more effective in performance than a combination of semantically different regression and classification losses (45.0$\rightarrow$46.3). Second, the performance of the NLL loss with a mixture model is relatively low because there is no means to remove duplicate bounding boxes (35.6). Third, the MCM loss to remove duplicates significantly improves the AP (35.6$\rightarrow$46.9). Fourth, the AP is slightly increased even if the matching method for the MCM loss is changed to a simple MAX function (46.9$\rightarrow$47.0). Also, D-RMM significantly outperforms Sparse R-CNN.

\begin{figure}[t]
\includegraphics[width=1.0\linewidth]{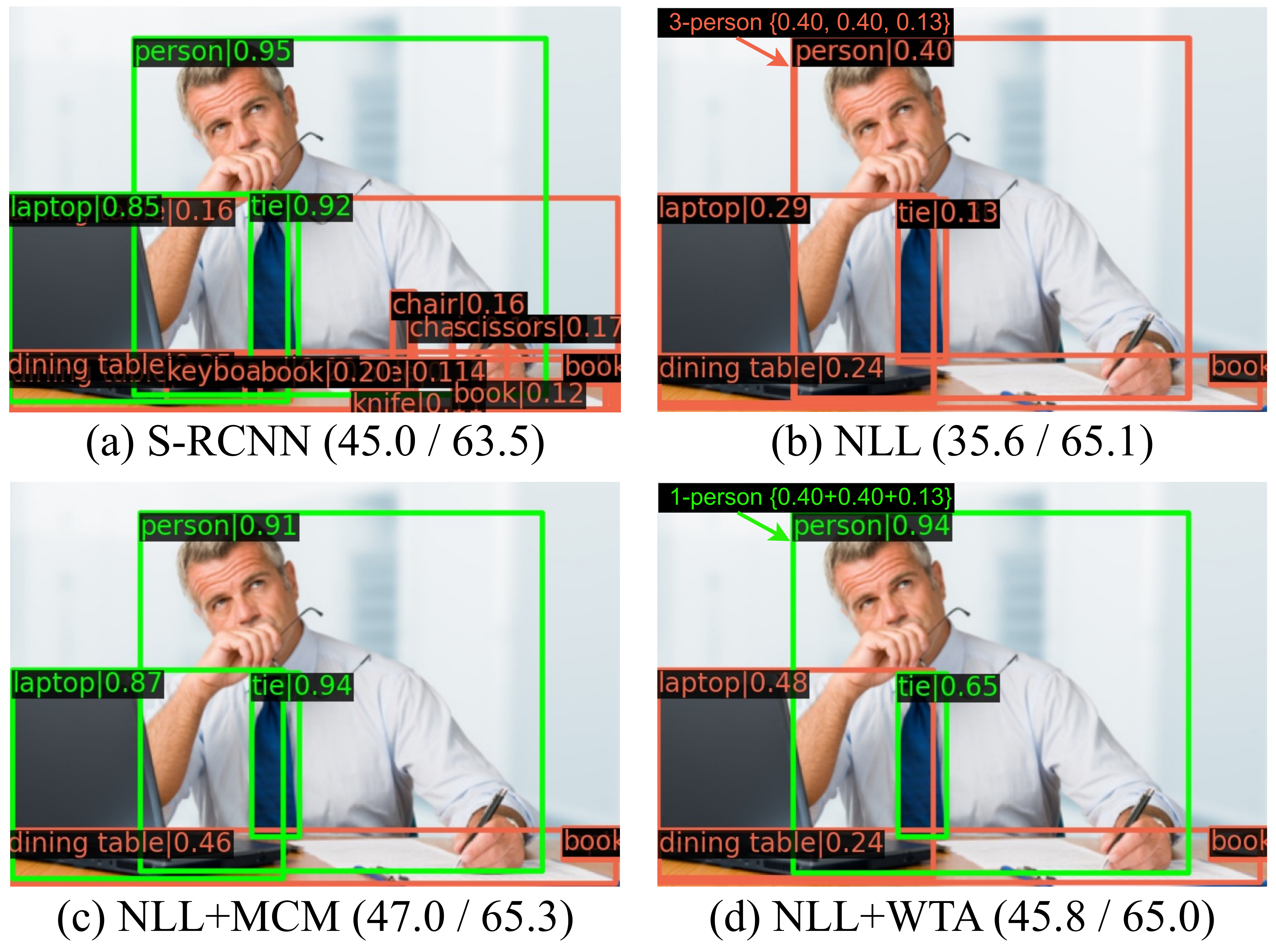}
\vspace{-3.0mm}
\caption{Qualitative result. Red/Green boxes indicate confidence 0.1-0.5 and 0.5-1.0, respectively. `/' separates AP/AR} 
\label{fig:vis_ablation}
\end{figure}

\begin{figure}[t]
  \centering
  \includegraphics[width=1.0\linewidth]{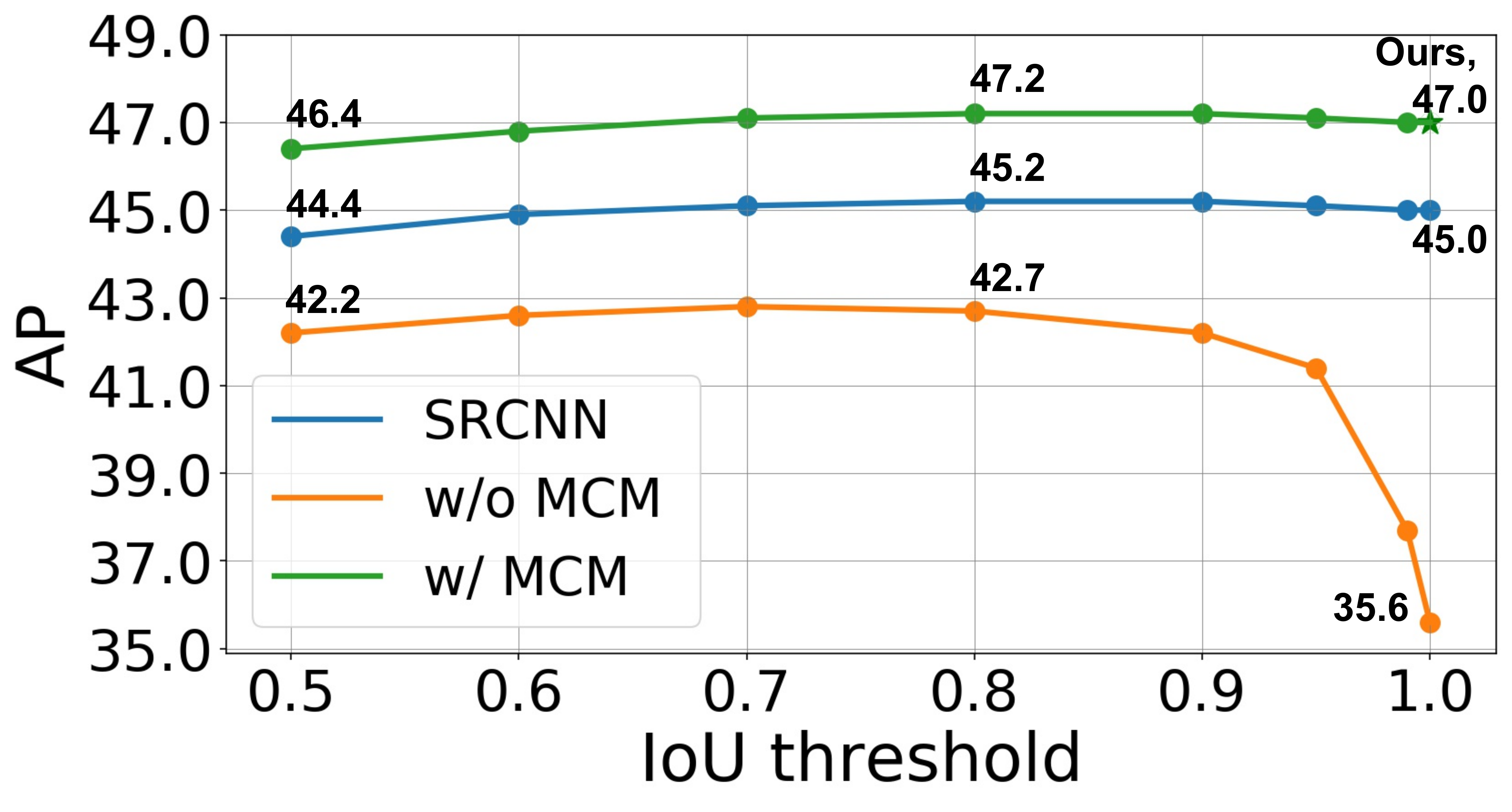}
  \vspace{-3.0mm}
  \caption{NMS result. The IoU threshold=1.0 means that NMS is not performed.}
  \label{fig:NMS_AP}
\end{figure}

\vspace{0mm}
\noindent \textbf{Visualization.} \quad Figure \ref{fig:vis_ablation} (a), (b), (c) visualize the 1st, 3rd, and last rows of\ Table \ref{tab:from_srcnn_to_ours}. As shown in (a) and (c), the inference results with confidence scores of 0.5 or higher are quite similar. However, with the confidence scores below 0.5, S-RCNN results are rather noisy. Although (b) seems to show comparable results to (c), the confidence scores of (b) tend to be lower than those of (c), and many overlapping boxes exist. For example, in (b), there are 3-overlapping boxes with low confidence scores of \{0.40, 0.40, 0.13\}. Within the mixture model framework, the likelihood of (b) might be similar to (c). It will be further discussed in Section \ref{sec:lms_loss_and_deduplication}. More examples are in Appendix \ref{appendix:sec:more_visualization}.

As shown in the figure, (b) has many duplicates. Interestingly, (b) and (c) has a large AP gap, but AR is similar. We conjectured the MCM loss contributed significantly to deduplication. Therefore, we conducted related experiments.

\begin{table}[t]
\vspace{-3mm}
\caption{Handicraft method for merging confidences by applying the winner-take-all strategy (WTA).}
\footnotesize
\vspace{0.5mm}
\renewcommand*{\arraystretch}{1.10}
\begin{center}
\begin{tabular}{L{4.85cm}|C{0.80cm}|C{0.80cm}}
\hline
Method & AP & AR \\
\hline\hline
(\uppercase\expandafter{\romannumeral1})\, $\mathcal{L}_{NLL}$ with Mixture model & 35.6 & 65.1 \\
(\uppercase\expandafter{\romannumeral2})\, (\uppercase\expandafter{\romannumeral1})+NMS & 42.7 & 65.0 \\ 
(\uppercase\expandafter{\romannumeral3})\, (\uppercase\expandafter{\romannumeral1})+WTA & 45.8 & 65.0 \\\hline
(\uppercase\expandafter{\romannumeral4})\, $\mathcal{L}_{NLL}$ only for MAX & 45.7 & 63.7 \\ \hline
(\uppercase\expandafter{\romannumeral5})\, $\mathcal{L}_{NLL}$+$\mathcal{L}_{MCM}$ & 47.0 & 65.3 \\
\hline
\end{tabular}
\label{tab:calib_disperse}
\vspace{-3.5mm}
\end{center}
\end{table}

\begin{table}[h]
\caption{Validation loss. $exp(-\mathcal{L}_{MCM})$ is $\max(g_i|X)/ p(g_i|X)$, which means the ratio of maximum mixture component to likelihood for each instance.}
\vspace{1.2mm}
\footnotesize
\begin{center}
\renewcommand*{\arraystretch}{1.05}
\begin{tabular}{C{2.5cm}|C{1.9cm}C{1.9cm}}
\hline
 & $w/o \ \mathcal{L}_{MCM}$ & $w/ \ \mathcal{L}_{MCM}$  \\ 
\hline
\hline
 $\mathcal{L}_{NLL}$ & 15.374 & 15.391  \\
$exp(-\mathcal{L}_{MCM})$  & 0.590 & 0.892  \\
\hline
\end{tabular}
\label{tab:val_loss_main}
\vspace{-2.0mm}
\end{center}
\end{table}

\vspace{0mm}
\subsection{Analysis of the MCM loss and deduplication}
\label{sec:lms_loss_and_deduplication}

\vspace{0mm}
\noindent \textbf{NMS.} \quad Figure \ref{fig:NMS_AP} is the result of applying NMS post-processing, although the models do not actually have the NMS post-processing at inference time. The MCM loss eliminates duplication as effectively as S-RCNN, showing little performance change even after applying NMS. The NMS with thresholds $\in \{0.7, 0.8, 0.9\}$ achieves a slight gain of performance for both S-RCNN and ours by eliminating a few overlapping boxes still remaining. However, in the case of the model without the MCM loss, it achieves a significant AP improvement through the NMS, which implies that a lot of duplicates exist. It indicates that the MCM loss plays a key role for the deduplication without the NMS post-processing. 

Table \ref{tab:val_loss_main} reports the validation loss of models with or without $L_{MCM}$. Obviously, there was a large difference in the MCM loss, but the NLL loss was similar. In other words, without $L_{MCM}$, the performance is low because duplication cannot be avoided, and the failure to estimate the density is not the cause of the low performance. Detailed experimental results are in Appendix \ref{appendix:sec:valid_los}.

\vspace{0mm}
\noindent \textbf{Winner-Take-All method for the dispersed confidences.} \quad In Figure \ref{fig:NMS_AP}, NMS improved the AP of the model trained only with NLL from 35.6\% to 42.7\%. However, it still does not reach 47.0\% of the model with the MCM loss. In order to reduce the AP gap, we applied a handicraft process called winner-take-all strategy (WTA), which is a way that the confidences of the boxes removed by NMS were added to those of the remaining boxes (Figure \ref{fig:vis_ablation} \subref{fig:sub_abl_wtl}). In Table \ref{tab:calib_disperse}, we compared (\uppercase\expandafter{\romannumeral3}) and (\uppercase\expandafter{\romannumeral4}) to see the effect of WTA, and there is an AP improvement from 42.7\% to 45.8\%. However, it was still below 47.0\% AP of (\uppercase\expandafter{\romannumeral5}). Rather than learning only with the NLL loss and merging it with manual rules, it is more effective to learn with the MCM loss to predict without overlapping. Furthermore, the MCM loss is straightforward and efficient in that there is no need for additional post-processing in the inference phase.

\subsection{Reliabilty and deduplication.}
\begin{figure}[t!]
    \centering
    \includegraphics[width=0.82\linewidth]{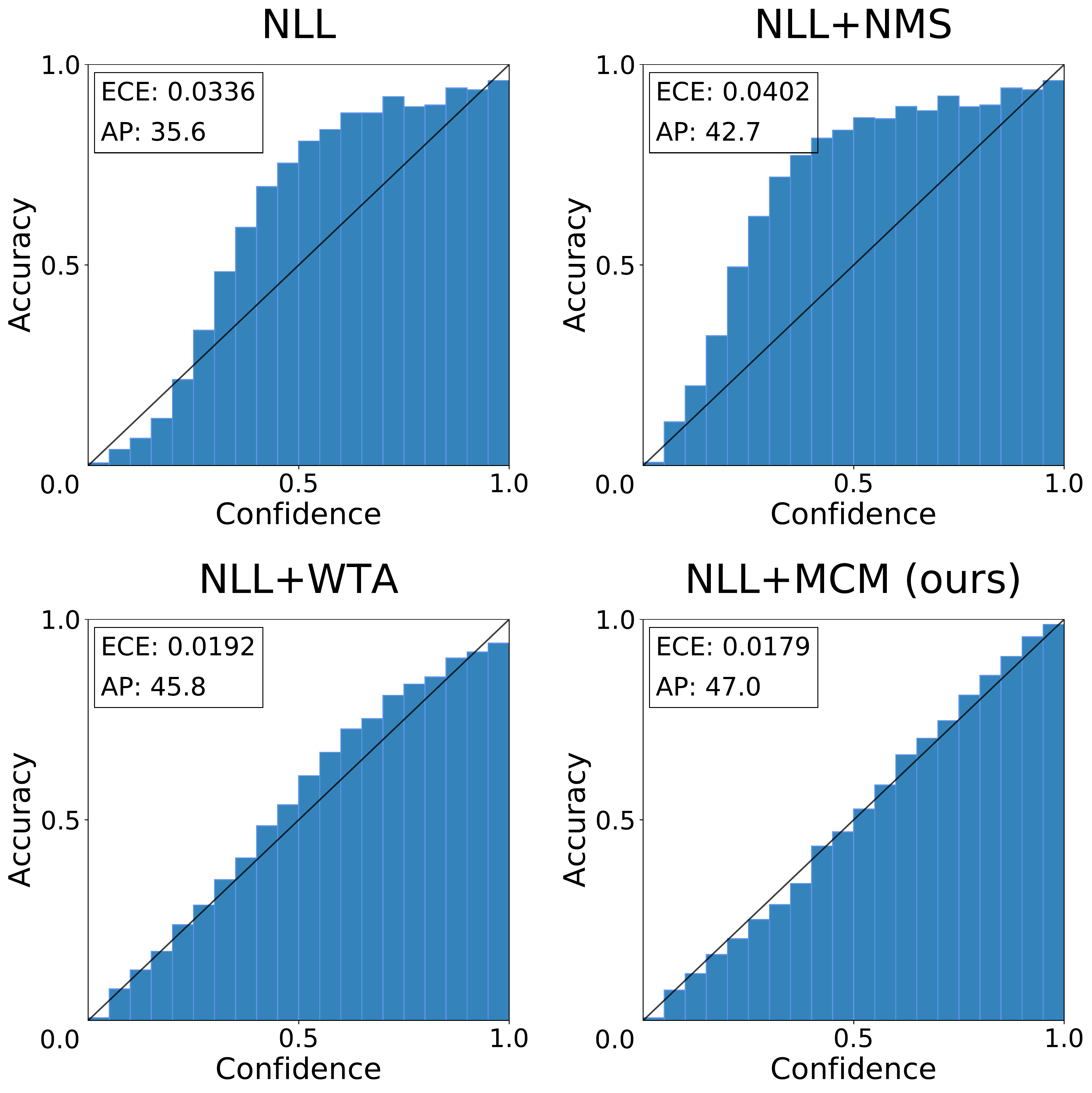}
    \vspace{-2.0mm}
    \caption{Confidence score and accuracy for model reliability with Expected Confidence Error (ECE) and AP.} 
    \vspace{0mm}
    \label{fig:ece_calibration}
\end{figure}

Figure \ref{fig:ece_calibration} shows the relationship between deduplication method and confidence reliability through confidence score and Expected Calibration Error (ECE), which calculates the difference between the confidence score and accuracy of each bin \cite{guo2017calibration}. ECE is defined as,
\begin{equation} 
ECE = \sum_{m} {|B_m| \over n}  \left| acc(B_m) - conf(B_m) \right|,
\end{equation}
where $B_m$ is $m$-th bin, $|B_m|$ is the $m$-th bin size, and $n$ is the total number of predictions. 

Although the NLL model is statistically trained, it is generally under-confident because the confidences are dispersed to multiple redundancies. Since the NMS removed duplicates, each bin's accuracy increased, but the ECE deteriorated. WTA, which was also influential in improving AP, was also effective in ECE. As the confidence of duplicates gathered together, the under-confidence effect was resolved similarly to ours.

Since D-RMM has an overall statistical training pipeline, including deduplication, confidence reliability is high. Besides, the model learned the mechanics of the WTA during the training process. Therefore, our model outperforms the WTA model for both AP and ECE and does not require a separate post-processing like WTA. From `NLL' to `NLL $+$ MCM', AP increases while ECE decreases.

\begin{figure}[t!]
    \centering
    \includegraphics[width=0.97\linewidth]{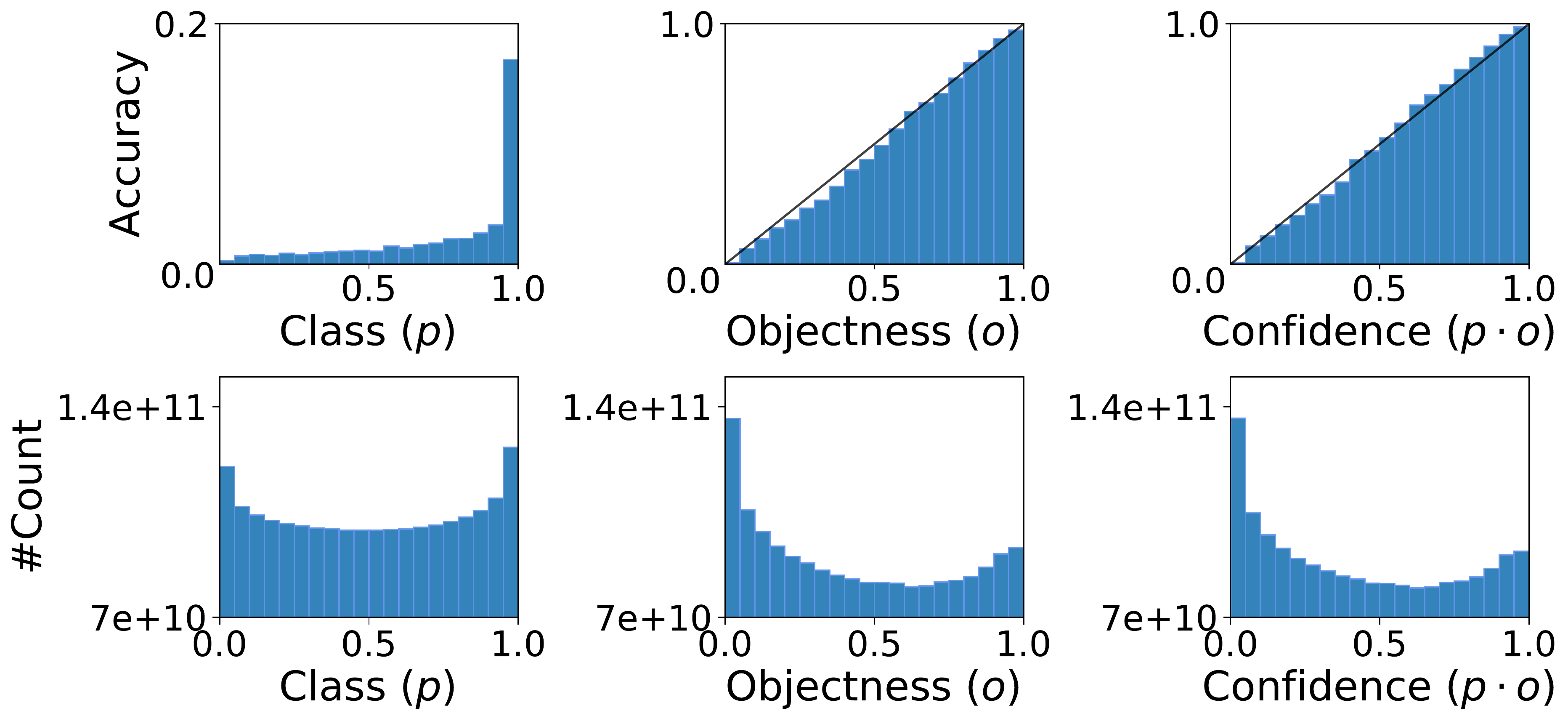}
    \vspace{-1.0mm}
    \caption{From left, illustrating histogram for class probability ($p$), objectness score ($o$), and confidence score ($p \cdot o$). Top: confidence histogram. Bottom: the number of predictions. The left and center histograms are plotted from the same bounding box predictions of the right.}
    \vspace{0mm}
    \label{fig:foreground_calib}
\end{figure}

In addition, we further analyzed the reliability of our model in Figure \ref{fig:foreground_calib}. If only class probability excluding the objectness score is considered (the left of Figure \ref{fig:foreground_calib}), there is a lot of duplication and it tends to be over-confident. On the other hand, the objectness score showed a similar histogram to the confidence score (the center and right of Figure \ref{fig:foreground_calib}).
In other words, objectness score plays a key role in model reliability.

\subsection{Comparison with others on COCO validation}
Table \ref{tab:val2017} shows the comparison with previous end-to-end detectors. Note that D-RMM has 300 proposal boxes, and we only evaluated the top-100 boxes. Compared to Sparse R-CNN and AdaMixer, from which we borrowed the structure, all of ours have significantly improved performance. Furthermore, D-RMM is effective not only in the ResNet backbone but also in the Swin Transformer-Tiny backbone.

Compared with the DETR variants, ours generally has higher performance. DN-Deformable-DETR, which denoising technique applied to DAB-Deformable-DETR, is similar to ours in R50 (48.6 and 48.4). We compared them by considering the speed in the next section.

\begin{figure}[t!]
    \centering
    \includegraphics[width=0.90\linewidth]{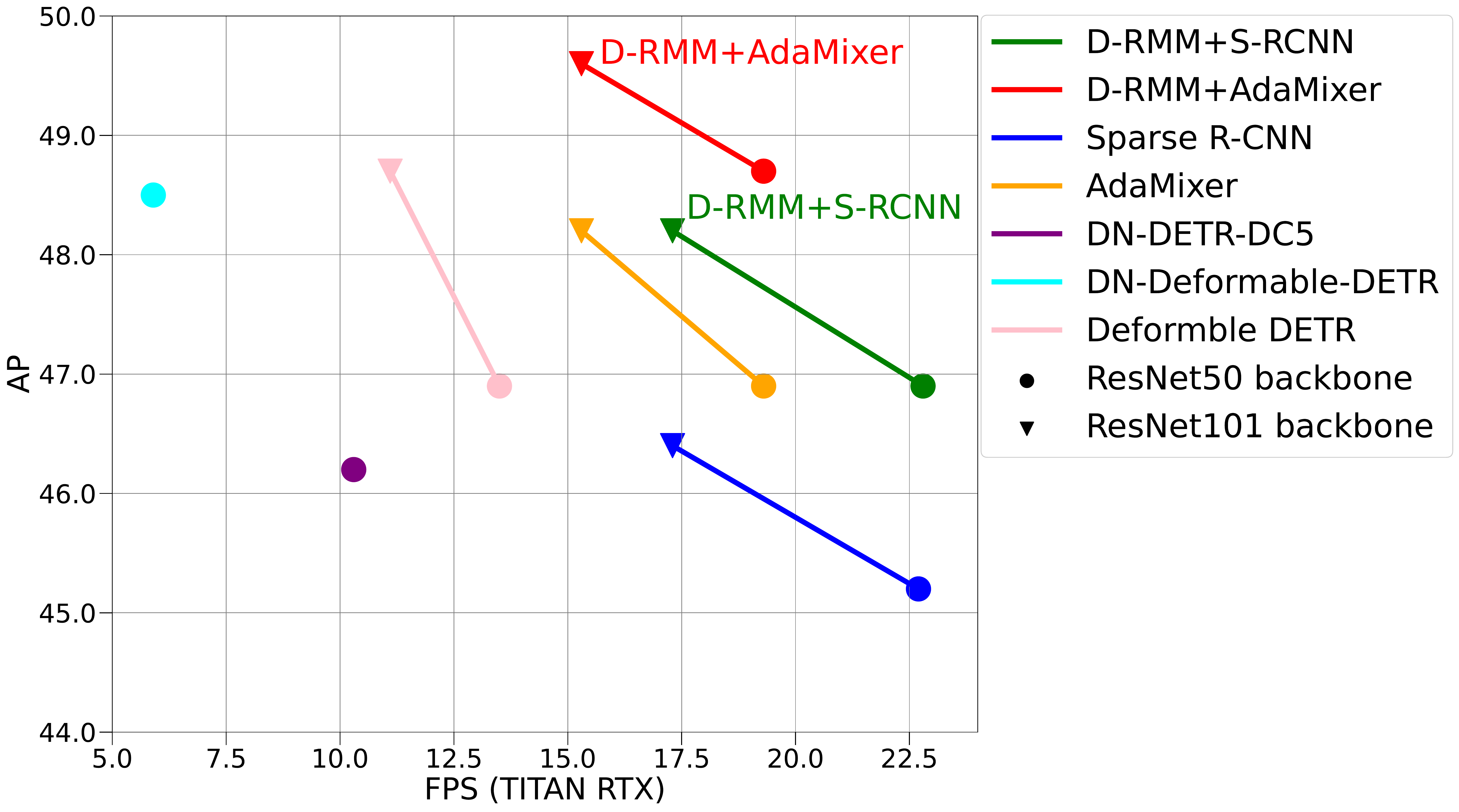}
    \vspace{-1mm}
    \caption{COCO test-dev results.} 
    \vspace{-1mm}
    \label{fig:comparision_on_testdev}
\end{figure}

\subsection{Comparison with others on COCO test-dev}
Figure \ref{fig:comparision_on_testdev} and Table \ref{tab:main_test_dev} compare ours with other state-of-the-art methods on COCO test-dev. Compared to Sparse R-CNN, ours have a significant AP improvement by +1.7\%p in ResNet50 and +1.8\%p in ResNet101. When we applied D-RMM to AdaMixer \cite{gao2022adamixer}, the performance improved over AdaMixer and showed the state-of-the-art performance. In the figure, ours stand out from the others regarding speed and performance.

\begin{table*}[hbt!]
\vspace{-1mm}
\caption{Comparision results with end-to-end Detectors on COCO validation. $\dagger$ uses 100 queries.}
\vspace{1.5mm}
\begin{center}
\renewcommand*{\arraystretch}{1.12}
\footnotesize
\begin{tabular}{L{5.6cm}|L{1.4cm}|C{1.0cm}|C{1.5cm}|C{0.75cm}C{0.75cm}|C{0.75cm}C{0.75cm}C{0.75cm}}
\hline
Method & Backbone & \#epochs &AP & AP$_{50}$ & AP$_{75}$ & AP$_{S}$ & AP$_{M}$ & AP$_{L}$ \\
\hline\hline
DETR \cite{carion2020detr} $\dagger$ & R50 & 500 & 42.0 & 62.4 & 44.2 &20.5 &45.8 & 61.1 \\
DETR \cite{carion2020detr} $\dagger$ & R101 & 500 &  43.5 & 63.8 & 46.4 & 21.9 & 48.0 & 61.8 \\
Conditional-DETR \cite{meng2021conditional} & R50 & 108 & 43.0 & 64.0 & 45.7 & 22.7 & 46.7 & 61.5 \\
Conditional-DETR \cite{meng2021conditional} & R101 & 108 &  44.5 & 65.6 & 47.5 & 23.6 & 48.4 & 63.6 \\
Deformable DETR \cite{zhu2020deformable} & R50 & 50 & 46.9 & 65.6 & 51.0 & 29.6 & 50.1 & 61.6 \\
Anchor DETR \cite{wang2022anchor} & R50 & 50 & 42.1 & 63.1 & 44.9 & 22.3 & 46.2 & 60.0 \\
Anchor DETR \cite{wang2022anchor} & R101 & 50 & 43.5 & 64.3 & 46.6 & 23.2 & 47.7 & 61.4 \\
DAB-Deformable-DETR  \cite{liu2022dab} & R50 & 50 & 46.9 & 66.0 & 50.8 & 30.1 & 50.4 & 62.5 \\
DN-DETR \cite{li2022dn}& R50 & 50 & 44.1 & 64.4 & 46.7 & 22.9 & 48.0 & 63.4 \\
DN-Deformable-DETR  \cite{li2022dn}& R50 & 50 & 48.6 & 67.4 & 52.7 & 31.0 & 52.0 &  63.7 \\
DN-DETR  \cite{li2022dn} & R101 & 50 & 45.2 & 65.5 & 48.3 & 24.1 & 49.1 & 65.1 \\
Sparse DETR \cite{roh2022sparse} & R50 & 50 &  46.3 & 66.0 & 50.1 & 29.0 & 49.5 & 60.8 \\
Sparse DETR \cite{roh2022sparse} & Swin-Tiny & 50 &  49.3 & \textbf{69.5} &53.3 & 32.0 &52.7 & 64.9 \\
ViDT \cite{song2022vidt} & Swin-Tiny & 150 & 47.2 & 66.7 & 51.3 & 28.4 & 50.2 & 64.7 \\
\hline
Sparse R-CNN \cite{sun2021sparse} & R50 & 36 & 45.0 & 64.1 & 49.0 & 27.8 & 47.6 & 59.7 \\
Sparse R-CNN \cite{sun2021sparse} & R101 & 36 &46.4 & 65.6 & 50.7 & 28.6 & 49.4 & 61.3 \\
Sparse R-CNN \cite{sun2021sparse} & Swin-Tiny & 36 &47.4 & 67.1 & 52.0 & 30.1 & 50.3 & 63.1 \\
AdaMixer \cite{gao2022adamixer} & R50 & 36 &47.0 & 66.0 & 51.1 & 30.1 & 50.2 & 61.8 \\
AdaMixer \cite{gao2022adamixer} & R101 & 36 &48.0 & 67.0 & 52.4 & 30.0 & 51.2 & 63.7 \\
AdaMixer \cite{gao2022adamixer} & Swin-Tiny & 36 &48.9 & 68.5 & 53.5 & 31.5 & 52.0 & 64.2 \\ \hline
D-RMM + Sparse R-CNN & R50 & 36 & 47.0 (+2.0) & 64.8 & 51.6 & 30.5 & 50.4 & 61.1 \\
D-RMM + Sparse R-CNN & R101 & 36 & 48.0 (+1.6) & 65.7 & 52.6 & 30.4 & 51.4 & 63.5 \\
D-RMM + Sparse R-CNN & Swin-Tiny & 36 & 49.9 (+2.5) & 68.1 &  55.1  & 32.1 &  53.6 & 64.6 \\
D-RMM + AdaMixer & R50 & 36 & 48.4 (+1.4) & 66.3 & 52.8 & 31.0 & 52.1 & 64.0 \\
D-RMM + AdaMixer & R101 & 36 & 49.2 (+1.2) & 67.3 & 53.5 & 31.4 & 53.1 & 65.4 \\
D-RMM + AdaMixer & Swin-Tiny & 36 & \textbf{50.7 (+1.8)} & 69.0 & \textbf{55.5} & \textbf{33.5} & \textbf{54.4} & \textbf{66.3} \\
\hline
\end{tabular}
\label{tab:val2017}
\end{center}
\vspace{-3mm}
\end{table*}

\begin{table*}[hbt!]
\caption{COCO test-dev result. FPS is measured on a single NVIDIA TITAN RTX with batch size 1, excluding the data loading time. $\ast$ excluded the denoising task when measuring FPS.}
\begin{center}
\renewcommand*{\arraystretch}{1.12}
\footnotesize
\begin{tabular}{L{5.6cm}|L{1.4cm}|C{1.5cm}|C{0.78cm}C{0.78cm}|C{0.78cm}C{0.78cm}C{0.78cm}|C{0.78cm}}
\hline
Method & Backbone & AP & AP$_{50}$ & AP$_{75}$ & AP$_{S}$ & AP$_{M}$ & AP$_{L}$ & FPS \\
\hline\hline
Deformable DETR \cite{zhu2020deformable} & R50 & 46.9 & 66.4 & 50.8 & 27.7 & 49.7 & 59.9 & 13.5 \\
Deformable DETR \cite{zhu2020deformable} & R101 & 48.7 & \textbf{68.1} & 52.9 & 29.1 & 51.5 & 62.0 & 11.1 \\
Dynamic DETR \cite{dai2021dynamic} & R50 & 47.2 & 65.9 & 51.1 & 28.6 & 49.3 & 59.1 & - \\
DN-DETR-DC5 \cite{li2022dn} $\ast$ & R50 & 46.2 & 66.4 & 49.6 & 24.5 & 49.3 & 62.7 & 10.3 \\
DN-Deformbale-DETR \cite{li2022dn} $\ast$ & R50 & 48.5 & 67.5 & 52.7 & 28.7 & 51.6 & 62.0 & 5.9 \\
\hline
Sparse R-CNN \cite{sun2021sparse} & R50 & 45.2 & 64.6 & 49.1 & 27.0 & 47.2 & 57.4 & 22.7 \\
Sparse R-CNN \cite{sun2021sparse} & R101 & 46.4 & 65.8 & 50.4 & 27.0 & 48.7 & 59.5 & 17.3 \\
AdaMixer \cite{gao2022adamixer} & R50 & 46.9 & 66.1 & 51.1 & 28.3 & 49.1 & 60.5 & 19.3 \\
AdaMixer \cite{gao2022adamixer} & R101 & 48.2 & 67.5 & 52.5 & 28.9 & 50.8 & 62.0 & 15.3 \\
\hline
D-RMM + Sparse R-CNN & R50 & 46.9 (+1.7) & 65.0 & 51.5 & 28.1 & 49.4 & 59.3 & 22.8 \\
D-RMM + Sparse R-CNN & R101 & 48.2 (+1.8) & 66.2 & 52.9 & 28.7 & 51.2 & 61.4 & 17.3 \\
D-RMM + AdaMixer & R50 & 48.7 (+1.8) & 66.8 & 53.1 & 29.0 & 51.6 & 62.1 & 19.3 \\
D-RMM + AdaMixer & R101 & \textbf{49.6 (+1.4)} & 67.7 & \textbf{54.1} & \textbf{29.8} & \textbf{52.5} & \textbf{64.0} & 15.3 \\
\hline
\end{tabular}
\vspace{0mm}
\label{tab:main_test_dev}
\end{center}
\vspace{0mm}
\end{table*}

\section{Conclusion}
Our D-RMM, an end-to-end object detector, has a simple pipeline because there is neither heuristic post-processing for duplication removal such as NMS at inference time nor heuristic box matching process at training time. The proposed MCM loss induces the detector to be trained to predict only one box with a high confidence score without duplication in each instance. Although D-RMM has the new pipeline and loss function that is not much deformed from the structure of Sparse R-CNN and AdaMixer, the performance is comparable to or better than other state-of-the-art methods. Furthermore, D-RMM has scalability because it has not changed the network structure. Not only can it be used with the latest backbone such as Swin Transformer, but it also has a large room to be applied to the improved network structure by changing only the output format and the loss function.

\section*{Acknowledgements}
The researchers at Seoul National University were funded by the Korean Government through the NRF grants 2021R1A2C3006659 and 2022R1A5A7026673 as well as IITP grant 2021-0-01343. 

\bibliography{icml2023}

\begin{thebibliography}{33}
\providecommand{\natexlab}[1]{#1}
\providecommand{\url}[1]{\texttt{#1}}
\expandafter\ifx\csname urlstyle\endcsname\relax
  \providecommand{\doi}[1]{doi: #1}\else
  \providecommand{\doi}{doi: \begingroup \urlstyle{rm}\Url}\fi

\bibitem[Carion et~al.(2020)Carion, Massa, Synnaeve, Usunier, Kirillov, and
  Zagoruyko]{carion2020detr}
Carion, N., Massa, F., Synnaeve, G., Usunier, N., Kirillov, A., and Zagoruyko,
  S.
\newblock End-to-end object detection with transformers.
\newblock In \emph{European conference on computer vision}, pp.\  213--229.
  Springer, 2020.

\bibitem[Chen et~al.(2019)Chen, Wang, Pang, Cao, Xiong, Li, Sun, Feng, Liu, Xu,
  Zhang, Cheng, Zhu, Cheng, Zhao, Li, Lu, Zhu, Wu, Dai, Wang, Shi, Ouyang, Loy,
  and Lin]{mmdetection}
Chen, K., Wang, J., Pang, J., Cao, Y., Xiong, Y., Li, X., Sun, S., Feng, W.,
  Liu, Z., Xu, J., Zhang, Z., Cheng, D., Zhu, C., Cheng, T., Zhao, Q., Li, B.,
  Lu, X., Zhu, R., Wu, Y., Dai, J., Wang, J., Shi, J., Ouyang, W., Loy, C.~C.,
  and Lin, D.
\newblock {MMDetection}: Open mmlab detection toolbox and benchmark.
\newblock \emph{arXiv preprint arXiv:1906.07155}, 2019.

\bibitem[Chen et~al.(2022)Chen, Chen, Zeng, and Wang]{chen2022group}
Chen, Q., Chen, X., Zeng, G., and Wang, J.
\newblock Group detr: Fast training convergence with decoupled one-to-many
  label assignment.
\newblock \emph{arXiv preprint arXiv:2207.13085}, 2022.

\bibitem[Dai et~al.(2021)Dai, Chen, Yang, Zhang, Yuan, and
  Zhang]{dai2021dynamic}
Dai, X., Chen, Y., Yang, J., Zhang, P., Yuan, L., and Zhang, L.
\newblock Dynamic detr: End-to-end object detection with dynamic attention.
\newblock In \emph{Proceedings of the IEEE/CVF International Conference on
  Computer Vision}, pp.\  2988--2997, 2021.

\bibitem[Deng et~al.(2009)Deng, Dong, Socher, Li, Li, and Fei-Fei]{imagenet}
Deng, J., Dong, W., Socher, R., Li, L.-J., Li, K., and Fei-Fei, L.
\newblock {ImageNet: A Large-Scale Hierarchical Image Database}.
\newblock In \emph{CVPR}, 2009.

\bibitem[Dugas et~al.(2000)Dugas, Bengio, B{\'e}lisle, Nadeau, and
  Garcia]{dugas2000softplus}
Dugas, C., Bengio, Y., B{\'e}lisle, F., Nadeau, C., and Garcia, R.
\newblock Incorporating second-order functional knowledge for better option
  pricing.
\newblock \emph{Advances in neural information processing systems}, 13, 2000.

\bibitem[Gao et~al.(2022)Gao, Wang, Han, and Guo]{gao2022adamixer}
Gao, Z., Wang, L., Han, B., and Guo, S.
\newblock Adamixer: A fast-converging query-based object detector.
\newblock In \emph{Proceedings of the IEEE/CVF Conference on Computer Vision
  and Pattern Recognition}, pp.\  5364--5373, 2022.

\bibitem[Guo et~al.(2017{\natexlab{a}})Guo, Pleiss, Sun, and
  Weinberger]{guo2017calibration}
Guo, C., Pleiss, G., Sun, Y., and Weinberger, K.~Q.
\newblock On calibration of modern neural networks.
\newblock In \emph{International conference on machine learning}, pp.\
  1321--1330. PMLR, 2017{\natexlab{a}}.

\bibitem[Guo et~al.(2017{\natexlab{b}})Guo, Pleiss, Sun, and
  Weinberger]{guo2017oncalibration}
Guo, C., Pleiss, G., Sun, Y., and Weinberger, K.~Q.
\newblock On calibration of modern neural networks.
\newblock In \emph{International conference on machine learning}, pp.\
  1321--1330. PMLR, 2017{\natexlab{b}}.

\bibitem[He et~al.(2016)He, Zhang, Ren, and Sun]{he2016resnet}
He, K., Zhang, X., Ren, S., and Sun, J.
\newblock Deep residual learning for image recognition.
\newblock In \emph{CVPR}, pp.\  770--778, 2016.

\bibitem[He et~al.(2019)He, Zhu, Wang, Savvides, and Zhang]{he2019bounding}
He, Y., Zhu, C., Wang, J., Savvides, M., and Zhang, X.
\newblock Bounding box regression with uncertainty for accurate object
  detection.
\newblock In \emph{Proceedings of the IEEE Conference on Computer Vision and
  Pattern Recognition}, pp.\  2888--2897, 2019.

\bibitem[Hu et~al.(2018)Hu, Gu, Zhang, Dai, and Wei]{hu2018relation}
Hu, H., Gu, J., Zhang, Z., Dai, J., and Wei, Y.
\newblock Relation networks for object detection.
\newblock In \emph{Proceedings of the IEEE conference on computer vision and
  pattern recognition}, pp.\  3588--3597, 2018.

\bibitem[Jia et~al.(2022)Jia, Yuan, He, Wu, Yu, Lin, Sun, Zhang, and
  Hu]{jia2022detrs}
Jia, D., Yuan, Y., He, H., Wu, X., Yu, H., Lin, W., Sun, L., Zhang, C., and Hu,
  H.
\newblock Detrs with hybrid matching.
\newblock \emph{arXiv preprint arXiv:2207.13080}, 2022.

\bibitem[Kuhn(1955)]{kuhn1955hungarian}
Kuhn, H.~W.
\newblock The hungarian method for the assignment problem.
\newblock \emph{Naval research logistics quarterly}, 2\penalty0 (1-2):\penalty0
  83--97, 1955.

\bibitem[Li et~al.(2022)Li, Zhang, Liu, Guo, Ni, and Zhang]{li2022dn}
Li, F., Zhang, H., Liu, S., Guo, J., Ni, L.~M., and Zhang, L.
\newblock Dn-detr: Accelerate detr training by introducing query denoising.
\newblock In \emph{Proceedings of the IEEE/CVF Conference on Computer Vision
  and Pattern Recognition}, pp.\  13619--13627, 2022.

\bibitem[Lin et~al.(2014)Lin, Maire, Belongie, Hays, Perona, Ramanan,
  Doll{\'a}r, and Zitnick]{lin2014microsoft}
Lin, T.-Y., Maire, M., Belongie, S., Hays, J., Perona, P., Ramanan, D.,
  Doll{\'a}r, P., and Zitnick, C.~L.
\newblock Microsoft coco: Common objects in context.
\newblock In \emph{European conference on computer vision}, pp.\  740--755.
  Springer, 2014.

\bibitem[Lin et~al.(2017{\natexlab{a}})Lin, Doll{\'a}r, Girshick, He,
  Hariharan, and Belongie]{lin2017feature}
Lin, T.-Y., Doll{\'a}r, P., Girshick, R., He, K., Hariharan, B., and Belongie,
  S.
\newblock Feature pyramid networks for object detection.
\newblock In \emph{Proceedings of the IEEE conference on computer vision and
  pattern recognition}, pp.\  2117--2125, 2017{\natexlab{a}}.

\bibitem[Lin et~al.(2017{\natexlab{b}})Lin, Goyal, Girshick, He, and
  Doll{\'a}r]{lin2017focal}
Lin, T.-Y., Goyal, P., Girshick, R., He, K., and Doll{\'a}r, P.
\newblock Focal loss for dense object detection.
\newblock In \emph{Proceedings of the IEEE international conference on computer
  vision}, pp.\  2980--2988, 2017{\natexlab{b}}.

\bibitem[Liu et~al.(2022)Liu, Li, Zhang, Yang, Qi, Su, Zhu, and
  Zhang]{liu2022dab}
Liu, S., Li, F., Zhang, H., Yang, X., Qi, X., Su, H., Zhu, J., and Zhang, L.
\newblock Dab-detr: Dynamic anchor boxes are better queries for detr.
\newblock \emph{arXiv preprint arXiv:2201.12329}, 2022.

\bibitem[Liu et~al.(2016)Liu, Anguelov, Erhan, Szegedy, Reed, Fu, and
  Berg]{liu2016ssd}
Liu, W., Anguelov, D., Erhan, D., Szegedy, C., Reed, S., Fu, C.-Y., and Berg,
  A.~C.
\newblock Ssd: Single shot multibox detector.
\newblock In \emph{European conference on computer vision}, pp.\  21--37.
  Springer, 2016.

\bibitem[Liu et~al.(2021)Liu, Lin, Cao, Hu, Wei, Zhang, Lin, and
  Guo]{liu2021swin}
Liu, Z., Lin, Y., Cao, Y., Hu, H., Wei, Y., Zhang, Z., Lin, S., and Guo, B.
\newblock Swin transformer: Hierarchical vision transformer using shifted
  windows.
\newblock In \emph{Proceedings of the IEEE/CVF International Conference on
  Computer Vision}, pp.\  10012--10022, 2021.

\bibitem[Loshchilov \& Hutter(2017)Loshchilov and
  Hutter]{loshchilov2017decoupled}
Loshchilov, I. and Hutter, F.
\newblock Decoupled weight decay regularization.
\newblock \emph{arXiv preprint arXiv:1711.05101}, 2017.

\bibitem[Meng et~al.(2021)Meng, Chen, Fan, Zeng, Li, Yuan, Sun, and
  Wang]{meng2021conditional}
Meng, D., Chen, X., Fan, Z., Zeng, G., Li, H., Yuan, Y., Sun, L., and Wang, J.
\newblock Conditional detr for fast training convergence.
\newblock In \emph{Proceedings of the IEEE/CVF International Conference on
  Computer Vision}, pp.\  3651--3660, 2021.

\bibitem[Peng et~al.(2018)Peng, Xiao, Li, Jiang, Zhang, Jia, Yu, and
  Sun]{peng2018megdet}
Peng, C., Xiao, T., Li, Z., Jiang, Y., Zhang, X., Jia, K., Yu, G., and Sun, J.
\newblock Megdet: A large mini-batch object detector.
\newblock In \emph{Proceedings of the IEEE conference on Computer Vision and
  Pattern Recognition}, pp.\  6181--6189, 2018.

\bibitem[Redmon et~al.(2016)Redmon, Divvala, Girshick, and
  Farhadi]{redmon2016yolo}
Redmon, J., Divvala, S., Girshick, R., and Farhadi, A.
\newblock You only look once: Unified, real-time object detection.
\newblock In \emph{Proceedings of the IEEE conference on computer vision and
  pattern recognition}, pp.\  779--788, 2016.

\bibitem[Ren et~al.(2015)Ren, He, Girshick, and Sun]{ren2015fasterRCNN}
Ren, S., He, K., Girshick, R., and Sun, J.
\newblock Faster r-cnn: Towards real-time object detection with region proposal
  networks.
\newblock In \emph{NIPS}, pp.\  91--99, 2015.

\bibitem[Rezatofighi et~al.(2019)Rezatofighi, Tsoi, Gwak, Sadeghian, Reid, and
  Savarese]{rezatofighi2019generalized}
Rezatofighi, H., Tsoi, N., Gwak, J., Sadeghian, A., Reid, I., and Savarese, S.
\newblock Generalized intersection over union: A metric and a loss for bounding
  box regression.
\newblock In \emph{Proceedings of the IEEE/CVF conference on computer vision
  and pattern recognition}, pp.\  658--666, 2019.

\bibitem[Roh et~al.(2022)Roh, Shin, Shin, and Kim]{roh2022sparse}
Roh, B., Shin, J., Shin, W., and Kim, S.
\newblock Sparse {DETR}: Efficient end-to-end object detection with learnable
  sparsity.
\newblock In \emph{International Conference on Learning Representations}, 2022.
\newblock URL \url{https://openreview.net/forum?id=RRGVCN8kjim}.

\bibitem[Song et~al.(2022)Song, Sun, Chun, Jampani, Han, Heo, Kim, and
  Yang]{song2022vidt}
Song, H., Sun, D., Chun, S., Jampani, V., Han, D., Heo, B., Kim, W., and Yang,
  M.-H.
\newblock Vi{DT}: An efficient and effective fully transformer-based object
  detector.
\newblock In \emph{International Conference on Learning Representations}, 2022.
\newblock URL \url{https://openreview.net/forum?id=w4cXZDDib1H}.

\bibitem[Sun et~al.(2021)Sun, Zhang, Jiang, Kong, Xu, Zhan, Tomizuka, Li, Yuan,
  Wang, et~al.]{sun2021sparse}
Sun, P., Zhang, R., Jiang, Y., Kong, T., Xu, C., Zhan, W., Tomizuka, M., Li,
  L., Yuan, Z., Wang, C., et~al.
\newblock Sparse r-cnn: End-to-end object detection with learnable proposals.
\newblock In \emph{Proceedings of the IEEE/CVF Conference on Computer Vision
  and Pattern Recognition}, pp.\  14454--14463, 2021.

\bibitem[Wang et~al.(2022)Wang, Zhang, Yang, and Sun]{wang2022anchor}
Wang, Y., Zhang, X., Yang, T., and Sun, J.
\newblock Anchor detr: Query design for transformer-based detector.
\newblock In \emph{Proceedings of the AAAI conference on artificial
  intelligence}, volume~36, pp.\  2567--2575, 2022.

\bibitem[Yoo et~al.(2021)Yoo, Lee, Chung, Seo, and Kwak]{Yoo_2021_ICCV}
Yoo, J., Lee, H., Chung, I., Seo, G., and Kwak, N.
\newblock Training multi-object detector by estimating bounding box
  distribution for input image.
\newblock In \emph{Proceedings of the IEEE/CVF International Conference on
  Computer Vision (ICCV)}, pp.\  3437--3446, October 2021.

\bibitem[Zhu et~al.(2020)Zhu, Su, Lu, Li, Wang, and Dai]{zhu2020deformable}
Zhu, X., Su, W., Lu, L., Li, B., Wang, X., and Dai, J.
\newblock Deformable detr: Deformable transformers for end-to-end object
  detection.
\newblock In \emph{International Conference on Learning Representations}, 2020.

\end{thebibliography}
\bibliographystyle{icml2022}

\newpage
\appendix
\onecolumn

\section{Training details} \label{appendix:sec:training_details}
In training based on Sparse R-CNN, the batch size is 16. The identical data augmentations used in Deformable DETR \cite{zhu2020deformable} are used for multi-scale training, where the input image size is 480 $\sim$ 800 with random crop and random horizontal flip. We use AdamW \cite{loshchilov2017decoupled} optimizer with a weight decay of 5e-5 and a gradient clipping with an L2 norm of 1.0. We adopt the training schedule of 36 epochs with an initial learning rate of 5e-5, divided by a factor of 10 at the 27th and 33rd epoch, respectively. In training based on AdaMixer, following the AdaMixer schedule, we changed the learning rate decay epoch from (27, 33) to (24, 33).

\section{Stop-gradient of the MCM loss} \label{appendix:sec:stop_grad}
Table \ref{subtab:loss_detach} shows the results when the stop-gradient scheme is applied to the MCM loss where the likelihood Eq. \ref{eq:mog} of the main paper is calculated with three elements: the mixing coefficient $\pi$, the Cauchy $\mathcal{F}$ and the categorical distribution $\mathcal{P}$. Note that there is no stop-gradient when calculating the NLL loss. It is the best when the stop-gradient is applied only to the $\mathcal{F}$. The back-propagation toward the $\mathcal{F}$ might make the box's coordinates deviate from the optimal to suppress the likelihoods of the overlapping boxes unless the stop-gradient is applied to the $\mathcal{F}$. Since it is an unintended phenomenon, applying stop-gradient toward the $\mathcal{F}$ is reasonable. However, there is no dramatic performance change with or without the stop-gradient.

\begin{figure}[h]
\begin{minipage}[c]{0.5\linewidth}
\centering
\footnotesize
\vspace{-3.0mm}
\captionsetup{width=.9\linewidth}
    \captionof{table}{Stop-gradient of likelihood of the MCM loss.}
    \vspace{+2.54mm}
    \renewcommand*{\arraystretch}{1.66}
    \begin{tabular}{C{0.15cm}|C{1.7cm}|C{2.15cm}|C{0.6cm}|C{0.6cm}|C{0.6cm}}
    \hline
    \multicolumn{3}{c|}{Stop-gradient} & \multicolumn{3}{c}{Metric} \\ 
    \hline
    $\pi$ & Cauchy ($\mathcal{F}$) & categorical ($\mathcal{P}$) & AP & AP$_{50}$ & AP$_{75}$ \\
    \hline
    - & - & -  & 46.8 & 64.7 & 51.3 \\
    - & stop & -  & \textbf{47.0} & \textbf{64.8} & \textbf{51.6}\\
    - & stop & stop & 46.7  & 64.5 & 51.3 \\
    \hline
    \end{tabular}
    \label{subtab:loss_detach}
\end{minipage}
\begin{minipage}[c]{0.5\linewidth}
\centering
\footnotesize
\vspace{-3.0mm}
\captionsetup{width=.9\linewidth}
    \captionof{table}{The MCM loss weight $\beta$. `/' separates AP/AP$_{50}$}
    \vspace{2.5mm}
    \renewcommand*{\arraystretch}{1.4}
    \begin{tabular}{c|c|c|c|c}
    \hline
    \multirow{2}{*}{\shortstack{MCM\\ weight }} & \multicolumn{4}{c}{The number of proposals}  \\
    \cline{2-5} 
    & 100 & 300 & 500 & 1000 \\
    \hline
    0.4 & 43.9/61.8 & 46.8/64.2 & 47.1/64.5 & 47.5/64.7 \\
    0.5 & \textbf{43.9}/\textbf{61.9} & \textbf{47.0}/\textbf{64.8} & 47.1/64.7 & 47.7/65.0 \\
    0.6 & 43.9/61.8 & 46.8/64.8 & \textbf{47.3}/\textbf{65.2} & \textbf{47.7}/65.2 \\
    0.7 & 43.5 61.6 & 46.7/64.7 & 47.2/65.1 & 47.6/\textbf{65.4} \\
    \hline
    \end{tabular}
    \label{subtab:num_roi_and_lms_weight}
\end{minipage}
\end{figure}

\section{Analysis of the number of Proposals}
\subsection{The MCM loss weight, $\beta$}  \label{appendix:sec:mcm_beta}
Table \ref{subtab:num_roi_and_lms_weight} is an experiment on the number of proposals and the weight of the MCM loss $\beta$. When the number of RoIs is 100 and 300, the AP is the highest at $\beta= 0.5$. And when it is 500 and 1000, the AP is the highest at $\beta = 0.6$. It is assumed that the MCM weight for deduplication had to be a little stronger because the higher the number of proposals (500, 1000), the more room for duplication. The MCM weight was fixed at 0.5 for simplification because the AP was not sensitively changed.

\subsection{Top-k ratio for proposals} 
\begin{figure}[t]
  \centering
  \begin{minipage}[c]{0.48\linewidth}
  \vspace{+2.5mm}
  \includegraphics[width=1.00\linewidth]{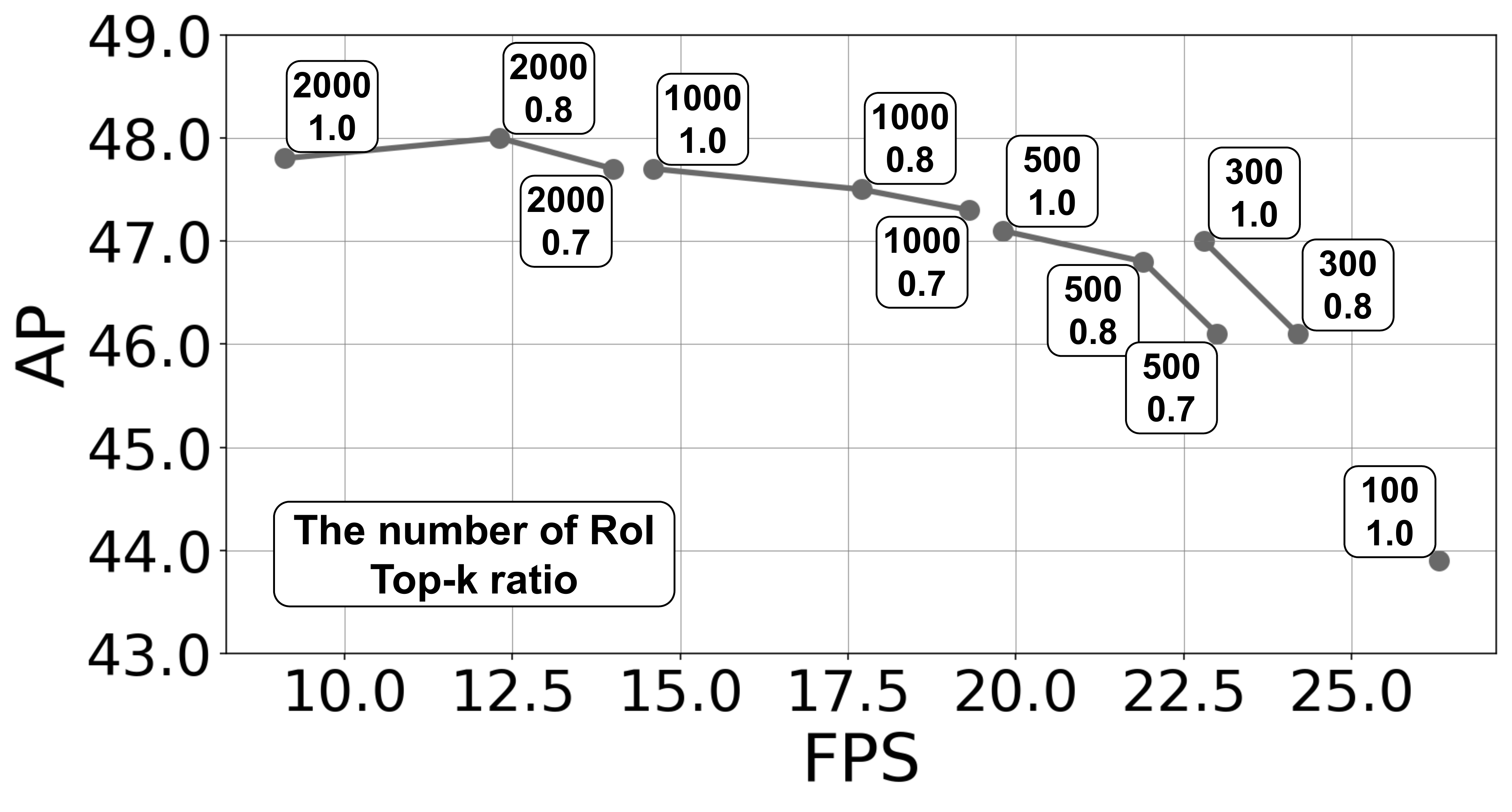}
  \vspace{-6.0mm}
  \caption{Top-k ratio for proposals. Top-k ratio denotes the percentage of remaining proposals after each stage.}
  \label{fig:graph_top_k}
  \end{minipage}
  \hfill
  \begin{minipage}[c]{0.48\linewidth}
    \includegraphics[width=1.00\linewidth]{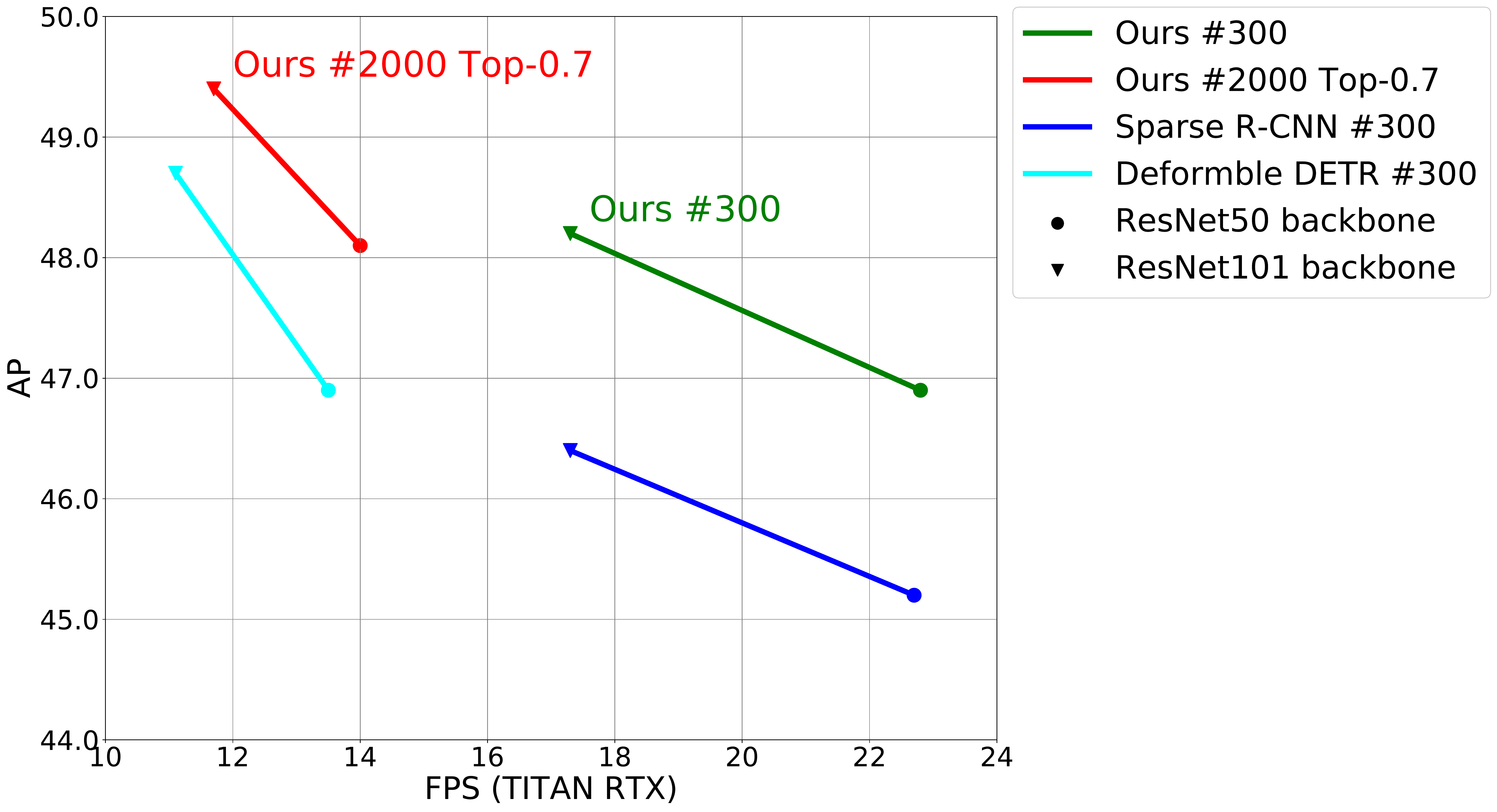}
    \caption{COCO test-dev results. \# is the number of proposals.}
    \label{fig:comparision_on_testdev-topk}
    \vspace{0mm}  
  \end{minipage}
\end{figure}

Figure \ref{fig:graph_top_k} shows how the performance changes when the number of proposals is reduced at each stage. The `top-k ratio' in the legend of the figure denotes the percentage of remaining proposals after every stage, and the number of remaining proposals in the final stage can be calculated by `\# of proposals $\times$ (top-k ratio)$^5$'. For example, if there are 2000 proposals in the 1st stage and the top-k ratio is 0.7, the number of final proposals is 2000 * $0.7^{5}\approx336$. As a result of the experiment, there is a trade-off between performance and speed. Furthermore, the more the number of proposals, the less the performance dropped even with a low top-k ratio. We conjecture that it is due to the enough number of final boxes remaining.

Figure \ref{fig:comparision_on_testdev-topk} shows the trade-off between speed and performance when top-k is applied. Then, D-RMM shows higher APs than Deformable DETR (+1.2\%p and +0.7\%p) at similar speeds.

\section{Validation loss} \label{appendix:sec:valid_los}
Table \ref{tab:val_loss_appen} shows the average validation loss of two models, which are trained with and without the MCM loss. 
This experiment demonstrates that the NLL and MCM loss contribute to the density estimation and deduplication, respectively.

The model without the MCM loss performs density estimation comparably well in terms of likelihood in that the NLL losses are similar in both cases. In other words, a lack of density estimation capability is not the root cause of the sharp drop of performance (AP 47.0\%$ \rightarrow$ 35.6\%). However, the two models show opposite tendencies in terms of $\exp(-\mathcal{L}_{MCM})=\max(g_i|X)/ p(g_i|X)$, which means the ratio of the maximum mixture component to likelihood. In the case of the model with the MCM loss, the ratio tends to increase as it goes through the stages. On the other hand, the model without the MCM loss has lower $exp(-\mathcal{L}_{MCM})$ for each stage than those of the model with the MCM loss. That is, the model without the MCM loss performs density estimation resulting in several overlapping predictions with low confidence in the final stage. As mentioned in Section 4.2 of the main paper, this result is consistent with that of Table 2 of the main paper in that the AP is significantly different depending on the presence of the MCM loss, whereas the AR is similar as 65.1\% and 65.3\% respectively.

\begin{table*}[h]
\caption{Validation loss, the NLL loss and the MCM loss. $exp(-\mathcal{L}_{MCM})$ is $\max(g_i|X)/ p(g_i|X)$, which means the ratio of maximum mixture component to likelihood for each instance.}
\begin{center}
\renewcommand*{\arraystretch}{1.12}
\begin{tabular}{C{0.7cm}|C{2.0cm}|C{2.0cm}|C{2.0cm}|C{2.0cm}|C{2.0cm}|C{2.0cm}}
\hline
\multirow{3}{*}{\shortstack{Stage}} & \multicolumn{3}{c|}{$\mathcal{L}_{NLL}$} & \multicolumn{3}{c}{$\exp(-\mathcal{L}_{MCM})$} \\ \cline{2-7}
& $w/ \ \mathcal{L}_{MCM}$ & $w/o \ \mathcal{L}_{MCM}$ & \multirow{2}{*}{$\cfrac{w/o \ \mathcal{L}_{MCM}}{w/ \ \mathcal{L}_{MCM}}$} & $w/ \ \mathcal{L}_{MCM}$ & $w/o \ \mathcal{L}_{MCM}$  & \multirow{2}{*}{$\cfrac{w/o \ \mathcal{L}_{MCM}}{w/ \ \mathcal{L}_{MCM}}$} \\[0.18ex]
& (AP 47.0) & (AP 35.6) &  & (AP 47.0) & (AP 35.6) & \\[0.18ex]
\hline 
1& 17.904 & 17.607 & 0.983 & 0.804 & 0.759 & 0.944 \\
2& 16.185 & 15.961 & 0.986 & 0.794 & 0.685 & 0.863 \\
3& 15.715 & 15.576 & 0.991 & 0.828 & 0.636 & 0.768 \\
4& 15.528 & 15.449 & 0.995 & 0.874 & 0.609 & 0.697 \\
5& 15.434 & 15.398 & 0.998 & 0.895 & 0.596 & 0.666 \\
6& 15.391 & 15.374 & 0.999 & 0.892 & 0.590 & 0.661 \\
\hline
\end{tabular}
\label{tab:val_loss_appen}
\end{center}
\end{table*}

\section{Limited supervision}
Current end-to-end methods learn ground truth information only from predictions selected by bipartite matching. The inherent ambiguity in the bounding box arising from occlusion, uncertain boundary, and inaccurate labeling \cite{he2019bounding} permits multiple plausible candidate bounding boxes for some objects. Hence, some objects have multiple possibilities for the bounding box representing them. However, due to bipartite matching, most end-to-end detectors learn one ground truth information through only one positive prediction. Therefore, other adjacent negative predictions are trained to be classified as background and cannot learn ground truth information at all, even if this prediction adequately represents another possible bounding box of the object.

\begin{figure}
\centering
\includegraphics[width=0.63\linewidth]{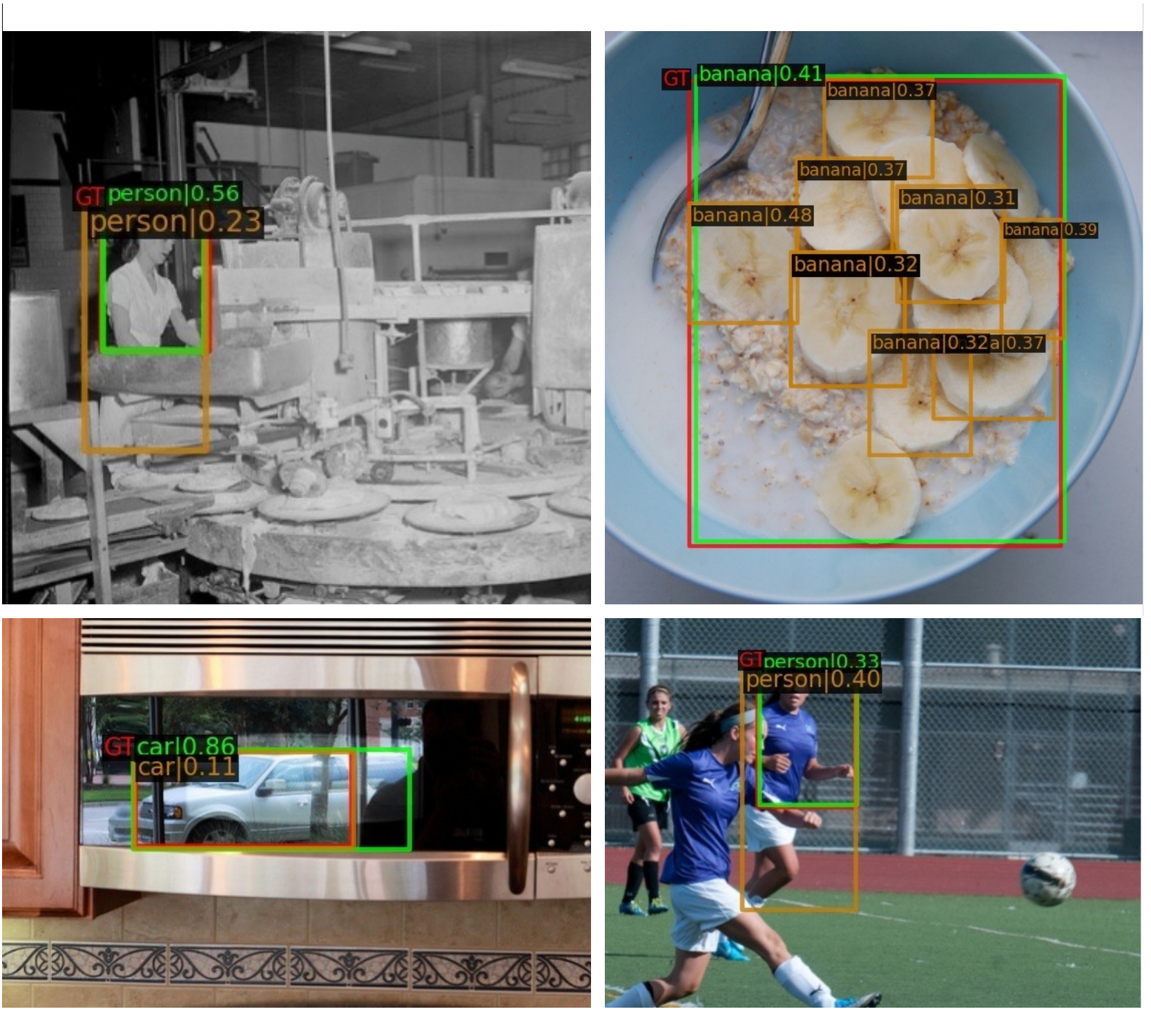}
\caption{Limited supervision. Red: GT, Green: Positive prediction, Orange: Negative prediction. Negative Predictions are plausible predictions, but the class confidences are trained as zero.} 
\label{fig:redundant}
\end{figure}

Figure \ref{fig:redundant} shows limited supervision examples. We take the banana class as an example (Figure \ref{fig:redundant} top right). The banana pieces showed high confidence, but each piece is assigned `negative' by the annotation rule, and the banana class confidences are learned as zero. That is, it learns that it is not a banana. On the other hand, the NLL loss can learn that various banana predictions are bananas in the categorical distribution of NLL (Eq. \ref{eq:loss_nll}). Whereas bipartite matching methods lose consistency when learning classification, ours is less prone to that with the NLL loss. The following question may arise. : The NLL loss is calculated together with multiple predictions. Then, isn't it possible that the predictions far from ground truth greatly influence the weight update, which can cause learning problems? The answer is in Section \ref{apen_sec:grad_our}.

\section{The number of positive assignments and interpretation in terms of the gradient}
Recently, studies on the relationship between the number of positive assignments and performance have been published, such as DN-DETR \cite{li2022dn}, Group-DETR \cite{chen2022group}, and Hybrid-DETR \cite{jia2022detrs}. We will interpret the NLL loss of our mixture model in terms of the number of positive assignments.

With the bipartite matcher, the number of positives in an image equals the number of ground truth boxes in that image. However, DN-DETR, Hybrid-DETR, and Group-DETR improved performance by increasing the number of assigned positives during training. In common, the three papers created additional Transformer Decoder branches used only during training and increased positives with the new query sets of the additional branches. DN-DETR increased the number of queries and positive assignments as a denoising task, and the other two papers performed additional bipartite matching on additional queries of new branches. (Computation cost and GPU usage significantly increase during training.)

The NLL loss of the mixture model can be viewed from the perspective of increasing positives with only one query set without additional branches. (Computation cost and GPU usage hardly increase during training.) For each ground truth, all predictions are learned as if they were positives. However, when calculating the gradient, the effect of each prediction is automatically adjusted. 

\subsection{Copy-Paste ground truth for additional positive assignment}
Table \ref{tab:num_pos} shows the correlation between performance and the number of positives of Sparse R-CNN. We increased positives by copying and pasting the ground truth. Note that we did not create any additional branches like Group-DETR in Table \ref{tab:num_pos}. We just copied and pasted the ground truths to simulate many-to-one matching. And, we applied NMS because it is many-to-one matching. 

As seen in the table, AP increased only by adequately increasing the positive. However, at some point, the increase in performance decreases, and even AP becomes similar to the default. On the other hand, Ours shows more significant performance improvements and no need for NMS, although the model learned all predictions as positives by the NLL loss.

\begin{table}[h]
\caption{The relationship between AP and the number of Copy-Paste of ground truth. The number of positive per ground truth is the same as the number of copies.}
\vspace{+2.54mm}
\renewcommand*{\arraystretch}{1.2}
\begin{center}
\begin{tabular}{c|c|c|c|c|c}
\hline
\multirow{2}{*}{NMS} & \multicolumn{5}{c}{The number of Copy-Paste per ground truth} \\
\cline{2-6}
& 1 & 2 & 3 & 4 & Ours (NLL+MCM) \\
\hline\hline
w/o &  45.0  &  28.5  & 21.1 & 17.4 & 47.0 \\
w/ &  45.2  &  46.1  & 45.6 & 45.3 & 47.2 \\
\hline
\end{tabular}
\label{tab:num_pos}
\vspace{0mm}
\end{center}
\end{table}

\subsection{Interpretation in terms of the gradient}  \label{appendix:sec:grad_analysis}
\subsubsection{Why does performance improvement decrease after Copy-Paste with bipartite matching?}
Let us assume three times Copy-Paste of the ground truths. The classification loss of positive boxes becomes Eq. (\ref{eq:analysis_grad_srcnn_a}) because the bipartite matching frameworks calculate the loss independently for each prediction. Then, the gradient becomes Eq. (\ref{eq:analysis_grad_srcnn_b}). Therefore, a prediction with an enormous loss significantly affects the total gradient.

\begin{subequations}
\label{eq:analysis_grad_srcnn}
\begin{flalign}
& \mathcal{L}_{cls} = -\log (p_i) -\log (p_j) -\log (p_k) \quad where, (i,j,k:assigned\quad index) \label{eq:analysis_grad_srcnn_a} \\
& \nabla_i \mathcal{L}_{cls} + \nabla_j \mathcal{L}_{cls} + \nabla_k \mathcal{L}_{cls} = - \frac{1}{p_i} - \frac{1}{p_j} - \frac{1}{p_k}
\label{eq:analysis_grad_srcnn_b}
\end{flalign}
\end{subequations}

This can negatively affect learning. The loss will induce relatively less well-fitting predictions to be learned strongly rather than inducing learning better to fit the predictions closest to the ground truth.

\begin{figure}[h]
\vspace{+1.0mm}
    \centering
    \includegraphics[width=0.95\linewidth]{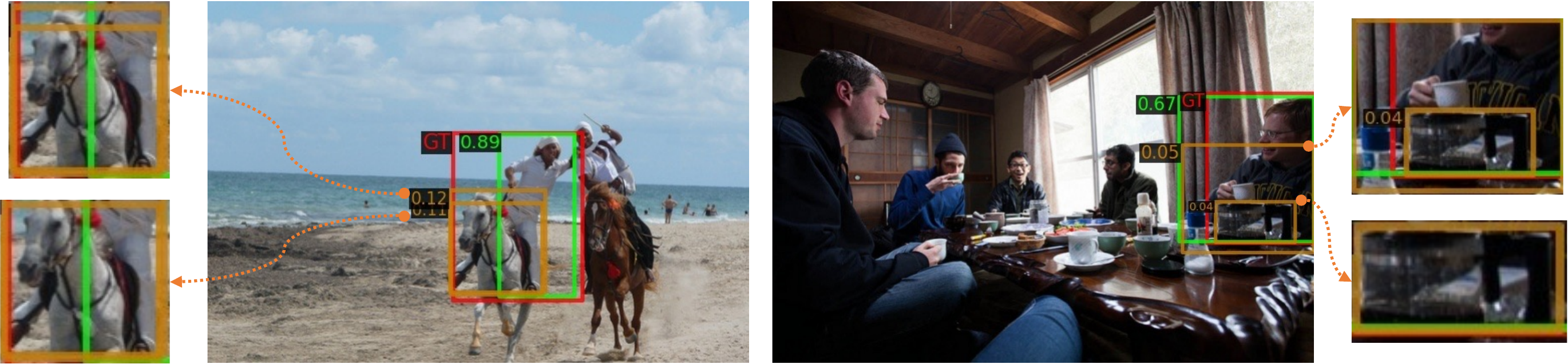}
    \caption{Examples of incorrect assignments when copied three times. For effective visualization, only one object was expressed. 
    The numbers next to the boxes are the confidence scores of the `person' class. Red: GT, Green: best assignment out of 3, Orange: the other two. Columns 2 and 3 are also learned as a `person' class.}
    \label{fig:misassign_example}
\end{figure}

Figure \ref{fig:misassign_example} shows examples. Even if it cannot be regarded as `person' and the confidence is relatively low, it is learned as `person' class. In this example, if the model learns with NLL, the effect of misassigned predictions on parameter update is reduced by the prediction that finds ground truth well. On the other hand, if learned with bipartite matching, the effect of incorrectly assigned prediction is relatively more significant by Eq. (\ref{eq:analysis_grad_srcnn_b}).

\subsubsection{Ours' NLL effectively increases performance even though it is all-to-one matching.} \label{apen_sec:grad_our}
Eq (\ref{eq:analysis_grad_ours}) is a simplified representation of Eq (5) of the main paper. $p_k$ is the likelihood of the $k$-th prediction. The gradient for the $k$-th prediction is Eq (\ref{eq:analysis_grad_ours_b}). 
\begin{subequations}
\label{eq:analysis_grad_ours}
\begin{flalign}
 & \mathcal{L}_{NLL} = -\log (p_1 + p_2 + ... +  p_K) \label{eq:analysis_grad_ours_a} \\
 & \nabla_k \mathcal{L}_{NLL} = - \frac{1}{(p_1 + p_2 + ... +  p_K)} \label{eq:analysis_grad_ours_b}
\end{flalign}
\end{subequations}

In Eq. (\ref{eq:analysis_grad_ours}), what has a more significant effect on the gradient is that the likelihood is more prominent, that is, closer to ground truth. Note that the likelihood of D-RMM is the joint of Categorical and Cauchy. In other words, if a prediction fits a certain ground truth well, other predictions that are wrong a lot have little effect on learning. 

\subsubsection{Multi-branch method}
Table \ref{tab:multi-branch} is the result of applying the Group-DETR method \cite{chen2022group} to Sparse R-CNN and AdaMixer. Note that Table \ref{tab:num_pos} is the result of many-to-one matching on just one branch without additional branches, and Table \ref{tab:multi-branch} has several branches and each branch performs one-to-one matching. Group-DETR has the effect of increasing the number of query groups, which independently performs bipartite matching. We experimented by adding 11-groups.

In Sparse R-CNN and AdaMixer, the AP increased significantly by additional groups but less than ours. Even if the number of positives is increased by applying the Group-DETR method, it is difficult to fundamentally avoid the hand-crafted assignment and inaccurate confidence score problem mentioned in Section \ref{sec:intro} because bipartite matching is performed in each group. Furthermore, due to additional groups, the GPU memory increases significantly.

\begin{table}[h]
\caption{Experiments utilizing Group-DETR \cite{chen2022group}. The additional group size is 11. The GPU memory is reported on 2-batch size per GPU using MMDet \cite{mmdetection}. FLOPs are measured for a sample of size 800x1280.}
\vspace{+2.54mm}
\footnotesize
\renewcommand*{\arraystretch}{1.1}
\begin{center}
\begin{tabular}{c|c|c|c|c|c}
\hline
 \multicolumn{2}{c|}{Method} & AP & Param (M) & Training GPU memory (GB)  & Inference FLOPs (G) \\
\hline\hline
 \multirow{3}{*}{Sparse R-CNN} & default &  45.0   & 106.1  & 7 & 165.5 \\
  & w/ Group &  45.9  & 107.0 & 21 & 165.5 \\ 
  & w/ ours &  47.0  &  106.1  & 7 &  165.5 \\ \hline
\multirow{3}{*}{AdaMixer} & default &  46.6 & 134.6  & 9 & 124.6 \\
  & w/ Group &  47.4  & 135.5 & 35 & 124.6 \\ 
  & w/ ours &  48.4  &  134.6 & 9 & 124.6 \\
\hline
\end{tabular}
\label{tab:multi-branch}
\vspace{0mm}
\end{center}
\end{table}

\section{Discussion for objective function and confidence} \label{appendix:sec:reliability}
We will discuss the reliability in terms of the cross-entropy loss (DETR) and the focal loss (Deformable DETR, Sparse R-CNN). We describe in Section \ref{sec:intro} that it is a key for reliability that the entire pipeline has a probabilistic point of view. In DETR, the classification with cross-entropy has a probabilistic basis, but bipartite matching and regression are not. Since cross-entropy is calculated based on hand-crafted bipartite matching, classification is not trained from the probabilistic point of view.

The focal loss is relatively well-calibrated compared to the cross-entropy. The focal loss adjusts the loss scale according to confidence; a weak loss for high confidence (easy example) and a substantial loss for low confidence (hard example). Therefore, the high confidence prediction is learned relatively weakly, and the low confidence is learned relatively strongly. So it becomes an S shape histogram. The low confidence area becomes overconfident, and the high confidence area becomes underconfident. Note that below the diagonal line is called overconfident, and the opposite is called underconfident. 

\newpage
\section{More examples for analysis} \label{appendix:sec:more_visualization}
\begin{figure*}[h!]
\centering
\includegraphics[width=0.77\linewidth]{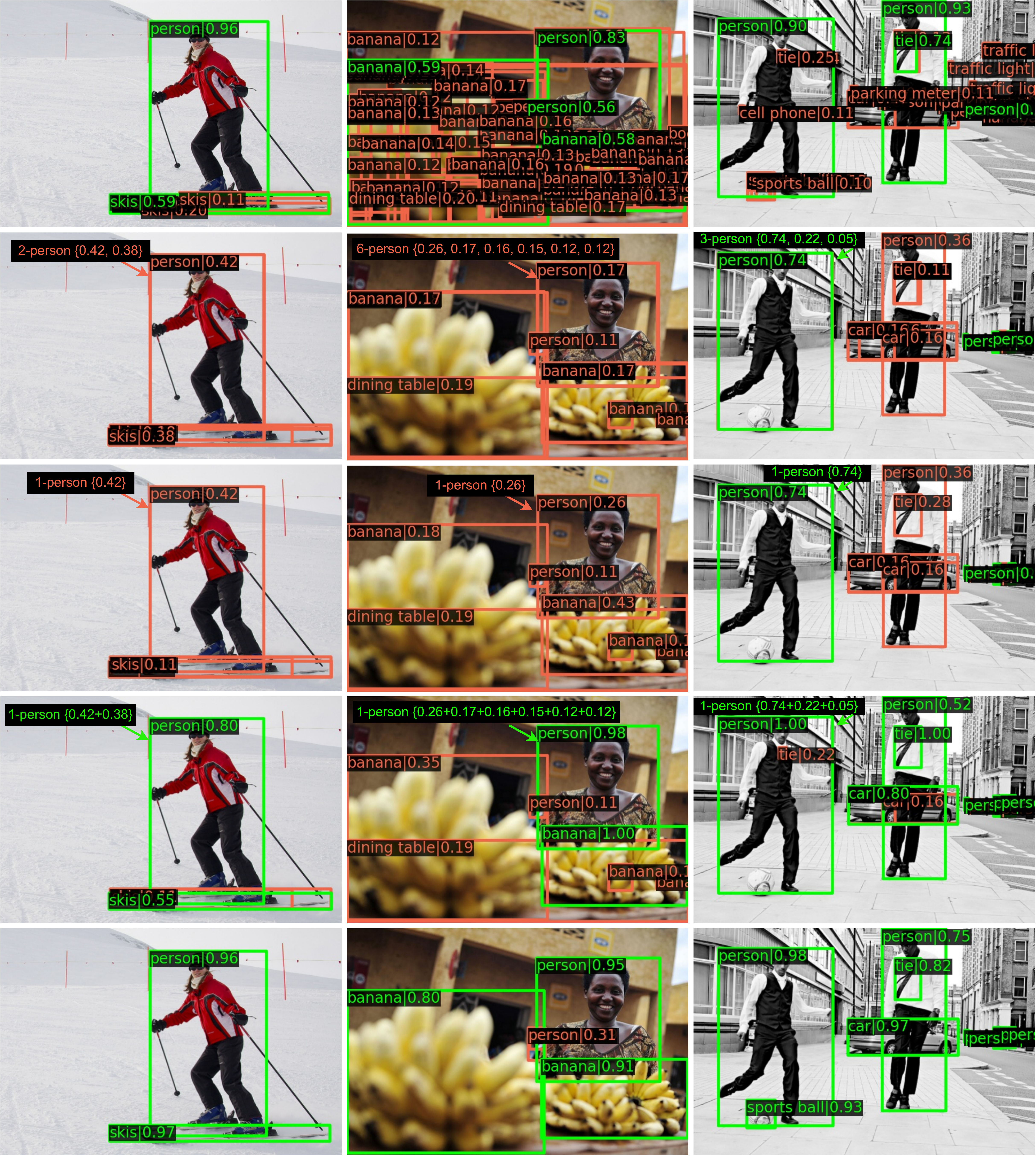}
\caption{Qualitative result. From above, 1st row:Sparse R-CNN (AP:45.0\%), 2nd row:NLL (AP:35.6\%), 3rd row:NLL+NMS (AP:42.7\%), 4th row:NLL+WTA (AP:45.8\%), 5th row:NLL+MCM (ours) (AP:47.0\%). Red/Green colors indicate confidence 0.1-0.5 and 0.5-1.0, respectively.}
\label{supp_fig:vis_comp_disperse}
\end{figure*}

Figure \ref{fig:vis_ablation} of the main paper shows qualitative result of Sparse R-CNN, model for ablation study and D-RMM. Figure \ref{supp_fig:vis_comp_disperse} shows more examples for these models.

\end{document}